\documentclass[sigconf, 10pt]{acmart}

\AtBeginDocument{%
  }
\usepackage{pifont}
\usepackage{colortbl}
\usepackage{multirow}
\usepackage{url}

\usepackage{amsmath,amssymb,amsthm,amsfonts,bm}
\usepackage{colortbl} 
\usepackage{xcolor}
\usepackage{array}  
\setcopyright{acmcopyright}
\usepackage{twemojis}
\newtheorem{theorem}{Theorem}

\newtheorem{lemma}{Lemma}
\usepackage[ruled,linesnumbered]{algorithm2e}
\usepackage[export]{adjustbox}
\usepackage{multicol}
\usepackage{diagbox}
\usepackage{enumitem}
\SetKwRepeat{Do}{do}{while}
\setcopyright{None}
\usepackage{pifont}
\usepackage{caption}
\usepackage{subfigure}
\usepackage{graphicx}

\copyrightyear{2024} 
\acmYear{2024}
\setcopyright{acmlicensed}\acmConference[ACM MobiCom '24]{The 30th Annual
International Conference on Mobile Computing and Networking}{November 18--22,
2024}{Washington D.C., DC, USA}
\acmBooktitle{The 30th Annual International Conference on Mobile Computing and
Networking (ACM MobiCom '24), November 18--22, 2024, Washington D.C., DC, USA}
\acmDOI{10.1145/3636534.3690701}
\acmISBN{979-8-4007-0489-5/24/11}

\begin{document}

\title[Cloud-assisted Data Enrichment for On-Device Continual Learning]{Delta: A Cloud-assisted Data Enrichment Framework for On-Device Continual Learning}

\author{Chen Gong}
\affiliation{
    \institution{Shanghai Jiao Tong University}
    \city{}
    \country{}
}
\author{Zhenzhe Zheng}
\affiliation{
    \institution{Shanghai Jiao Tong University}
    \country{}
    \city{}
    \authornote{Zhenzhe Zheng is the corresponding author.}
}
\author{Fan Wu}
\affiliation{
    \institution{Shanghai Jiao Tong University}
    \city{}
    \country{}
}
\author{Xiaofeng Jia}
\affiliation{
    \institution{Beijing Big Data Centre}
    \country{}
    \city{}
}
\author{Guihai Chen}
\affiliation{
    \institution{Shanghai Jiao Tong University}
    \country{}
    \city{}
}


\begin{abstract}
In modern mobile applications, users frequently encounter various new contexts, necessitating on-device continual learning (CL) to ensure consistent model performance.
While existing research predominantly focused on developing lightweight CL frameworks, we identify that data scarcity is a critical bottleneck for on-device CL.
In this work, we explore the potential of leveraging abundant cloud-side data to enrich scarce on-device data, and propose a private, efficient and effective data enrichment framework {\ttfamily Delta}.
Specifically, {\ttfamily Delta} first introduces a directory dataset to decompose the data enrichment problem into device-side and cloud-side sub-problems without sharing sensitive data.
Next, {\ttfamily Delta} proposes a soft data matching strategy to effectively solve the device-side sub-problem with sparse user data, and an optimal data sampling scheme for cloud server to retrieve the most suitable dataset for enrichment with low computational complexity.
Further, {\ttfamily Delta} refines the data sampling scheme by jointly considering the impact of enriched data on both new and past contexts, mitigating the catastrophic forgetting issue from a new aspect.
Comprehensive experiments across four typical mobile computing tasks with varied data modalities demonstrate that {\ttfamily Delta} could enhance the overall model accuracy by an average of $15.1\%$, $12.4\%$, $1.1\%$ and $5.6\%$ for visual, IMU, audio and textual tasks compared with few-shot CL, and consistently reduce the communication costs by over $90\%$ compared to federated CL.
\end{abstract}



\maketitle
\section{Introduction}
\label{sec: introduction}

Machine learning (ML) models have become the indispensable components in modern mobile applications and services, such as image tagging in Google Smart Lens~\cite{google_lens}, speech recognition in Siri~\cite{siri}, text summarization and rewriting in Apple Intelligence~\cite{apple_intelligence} and etc. 
In a wide range of mobile applications, users encounter dynamic contexts in their daily lives and exhibit varying behaviors, leading to a non-stationary data distribution observed and collected by mobile devices. 
Consequently, on-device ML models are expected to evolve incrementally as new contextual data becomes available.
This evolution, known as \textit{continual learning (CL})~\cite{thrun1995lifelong, silver2013lifelong}, enables on-device ML models to gradually learn individual user preferences in different contexts and behaviors, and thus becoming more personalized and intelligent over time.

Unlike conventional ML built on the premise of learning static data distributions, CL involves learning from dynamic data distributions. A significant challenge in CL is balancing the model's learning plasticity (\textit{i.e.} ability to assimilate new knowledge from emerging context) and memory stability (\textit{i.e.} ability to preserve past knowledge from historical contexts). 
For cloud servers with abundant hardware and data resources, many CL approaches have been proposed to address this challenge, such as regularizing model parameter updates~\cite{kirkpatrick2017overcoming, zenke2017continual}, replaying historical data~\cite{chaudhry2019tiny, li2017learning, lopez2017gradient} and designing context-adaptive model architectures~\cite{mallya2018piggyback, serra2018overcoming, kwon2023tinytrain}.
For resource-constrained devices, previous research focused on optimizing the usage of limited hardware resources to facilitate the efficient on-device deployment of cloud-side CL solutions~\cite{hayes2022online, kwon2023lifelearner}, such as saving storage through data quantization~\cite{ravaglia2021tinyml, hersche2022constrained}, accelerating data loading via hierarchical memory management~\cite{lee2022carm, ma2023cost}, and speeding up computation by optimizing the allocation of hardware resources~\cite{leite2022resource, kudithipudi2023design}.

\textbf{Data Bottleneck on Mobile Devices.}
However, we identify that the scarce data resource on mobile devices is the key bottleneck for on-device CL.
\textit{First, data scarcity is a pervasive issue across various mobile applications.} For example, for image analysis applications, an average European citizen takes only $4.9$ photos daily~\cite{image_scarcity}. For virtual assistant applications, a mere $16\%$ of iPhone users reports using Siri several times a day~\cite{siri_scarcity}. 
\textit{Second, the utilization of data resources fundamentally determines the performance ceiling for on-device CL}, whereas the optimization of hardware resources only influences the efficiency with which this ceiling can be reached.
On one hand, limited data resources for a single context often results in the well-known issue of model overfitting~\cite{hawkins2004problem, ying2019overview}.
On the other hand, the inadequate data resources for both past and new contexts exacerbate the mutual interference between their learning processes, which impedes knowledge transfer for new context and deteriorates the model performance on past contexts, a phenomenon commonly referred to as catastrophic forgetting~\cite{french1999catastrophic, mccloskey1989catastrophic, kirkpatrick2017overcoming}.

\textbf{Limitation of Existing Work.}
To tackle the challenge of data scarcity for CL, \textit{few-shot CL} and \textit{federated CL} are two representative approaches to mitigate the issues of overfitting and catastrophic forgetting from the aspects of model initialization and training algorithms (elaborated in \S\ref{sec: limitation of existing ideas}).\\
\textit{(1) Few-shot CL}~\cite{tao2020few, mazumder2021few, shi2021overcoming} involves pre-training ML models on common contexts with extensive data to capture general knowledge, which can be transferred to new contexts through model initialization and transfer learning techniques. However, this approach is ineffective for on-device settings due to the unpredictability and diversity of upcoming user contexts.
\textit{(2) Federated CL}~\cite{yoon2021federated, dong2022federated} suggests leveraging a cloud server to periodically aggregate the local models trained on distributed devices, which mitigates the overfitting problem on a single device and enables knowledge transfer across multiple devices. However, the model performance and convergence rate of federated CL are sensitive to device participation rate and data heterogeneity across devices~\cite{li2022pyramidfl, shin2022fedbalancer, gong2024ode}, leading to high communication overhead and unstable training process for real-world applications. 

\textbf{Our Motivation.}
The data bottleneck of mobile devices coupled with the limitations of existing approaches motivate us to consider leveraging the abundant cloud-side data resources to enrich the sparse device-side data, fundamentally addressing the data scarcity problem. As we will elaborate in \S\ref{sec: limitation of existing ideas}, simply increasing the training data size from $10$ to $50$ can yield a $10\%$ improvement in model accuracy compared to the best few-shot CL approach, while incurring less than $5\%$ communication costs compared with federated CL. The feasibility of such a cloud-assisted data enrichment framework is underpinned by two key observations:
\textit{(1) Abundant cloud-side data resource.} 
Cloud servers typically possess extensive datasets sourced from various channels, such as public datasets released by organizations \textit{(e.g.} ImageNet~\cite{russakovsky2015imagenet}), open-source data crawled from the Internet webs (\textit{e.g.} Common Crawl~\cite{common_crawl}), crowdsourced data contributed by authorized mobile users (\textit{e.g.} DonateClient service of Huawei~\cite{huawei_datadonation} and learn from this app in Apple~\cite{apple_learn_from_app}).
\textit{(2) Similarities among user contexts and behaviors.}
Previous investigations have demonstrated that the preferences and behaviors of different mobile users in various contexts share similar patterns rather than being entirely unique~\cite{bao2012location, lv2013mining, gong2023store}. This indicates the existence of a cloud-side data-subset that exhibits a similar distribution with the device-side data, offering an opportunity to enhance on-device CL performance.

\textbf{Challenges.}
A feasible data enrichment framework for practical on-device CL needs to be \textit{private}, \textit{effective} and \textit{efficient}, which are challenging to be achieved simultaneously.

\noindent \textit{$\bullet$ Privacy vs. Efficiency.}
In contemporary mobile applications, user data stored on devices is subject to stringent privacy regulations like GDPR~\cite{gdpr}. However, to enrich device-side data with an optimal data-subset from cloud, one must either
upload raw user data to the cloud for precise similarity comparison~\cite{yan2022device},  
or download numerous data-subsets from cloud and conduct trial-and-error processes to identify the appropriate data-subset~\cite{chai2022selective}.
Therefore, achieving efficient data enrichment without violating user privacy is challenging.

\noindent \textit{$\bullet$ Effectiveness vs. Efficiency for New Context.}
Given the diverse sources of cloud-side data, a randomly selected data-subset is likely to deviate significantly from the device-side data distribution, thereby degrading the CL performance over personal contexts.
However, to identify the  data-subset with the highest data enrichment performance for on-device CL, the cloud server needs to evaluate an exponential number of candidate data-subsets from the vast cloud-side dataset, which introduces prohibitively high time complexity and computational burden. 
Consequently, simultaneously reaching high effectiveness and efficiency poses another challenge.

\noindent \textit{$\bullet$ Effectiveness for Both Past and New Contexts.}
As the data distributions of new contexts encountered by mobile users are dynamic, independently conducting data enrichment for each emerging context would compromise the on-device model's memory stability over past contexts, as the mutual interference among different contexts' learning processes can be escalated.
Additionally, there is a lack of theoretical analysis or insight into the correlation between the enriched data of new context and model performance over past contexts, which further complicates the data enrichment problem for CL.
Therefore, designing a data enrichment strategy that is effective for both new and past contexts is challenging.

\textbf{Our Design.}
We propose {\ttfamily Delta}, a cloud-assisted data enrichment framework designed for on-device CL with high privacy protection, efficiency and effectiveness.
First, we provide a generic formalization of the data enrichment problem for on-device CL, and analyze its practical challenges concerning user privacy and computation efficiency. 
Second, to mitigate privacy concerns, we propose the construction of a compact ``directory'' dataset for cloud-side data. This approach helps to decompose the original data enrichment problem into two sub-problems, which can be independently solved by mobile device and cloud server without necessitating the exchange of sensitive raw data.
Third, to achieve both efficient and effective data enrichment for each new context, we develop a soft data matching strategy to accurately solve the device-side sub-problem with sparse on-device data, and a theoretically optimal data sampling scheme for cloud-side data selection, which can be computed with a constant time complexity.
Fourth, to maintain high effectiveness across both new and past contexts, we theoretically analyze the impact of new context's enriched data on model performance over all contexts, and re-optimize cloud-side data sampling strategy from a holistic perspective.

\textbf{Contributions} of this work are summarized as follows:\\
{$\bullet$} We identify the data bottleneck in on-device CL for dynamic user contexts, and explore the potential of utilizing cloud-side abundant data to enrich device-side data.\\
{$\bullet$} We formalize the data enrichment problem for on-device CL and propose the first practical cloud-assisted data enrichment framework that simultaneously achieves privacy protection, effectiveness and efficiency.\\
{$\bullet$} We evaluate {\ttfamily Delta} across four typical mobile computing tasks with diverse data modalities and models, demonstrating its broad applicability and superior performance over baselines in overall accuracy and communication efficiency.

\section{Background and Motivation}
\label{sec: background}
\subsection{On-Device Continual Learning}
\label{sec: on-device continual learning}
\begin{figure}
    \centering
    \vspace{-0.2cm}
    \includegraphics[width=0.47\textwidth]{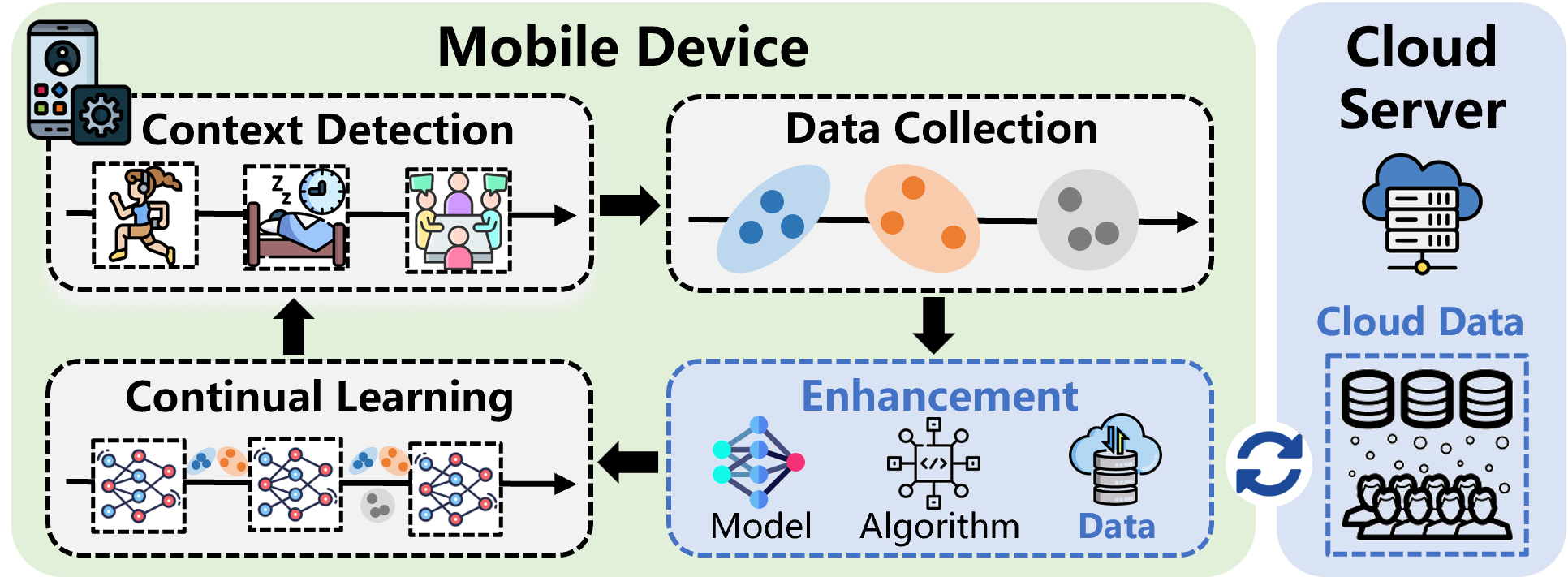}
    \vspace{-0.25cm}
    \caption{On-device continual learning pipeline.}
    \Description{On-device continual learning pipeline.}
    \vspace{-0.35cm}
    \label{fig: on-device CL pipeline}
\end{figure}

In mobile applications, users often encounter dynamic contexts and exhibit varying behaviors, leading to a non-stationary distribution of data collected by devices.
For example, mobile users can encounter unseen objects, weather conditions and digital corruptions in image analytics applications~\cite{bhardwaj2022ekya, khani2023recl}, 
experience new activities, physical conditions and device placements in human activity recognition (HAR) applications~\cite{kwon2021exploring}, 
or come across articles on various topics and in different languages in text analysis applications~\cite{ke2021achieving}.
These applications necessitate timely and accurate responses from on-device ML models to ensure high service quality, driving the need for on-device CL.
Figure \ref{fig: on-device CL pipeline} depicts the four stages a new context undergoes in on-device CL.

\noindent \textit{$\bullet$ Context Detection:}
When a new context is experienced by the user, it can be detected by mobile device through existing human-involved or automatic approaches~\cite{yin2008sensor, bhardwaj2022ekya, khani2023recl}. For example, in HAR application, the former approach would suggest users to confirm a new activity, whereas the latter would detect a shift in sensor data distribution~\cite{yin2008sensor}.

\noindent \textit{$\bullet$ Data Collection:}
For each new context, data samples following a new distribution are collected by mobile device as training data for the subsequent on-device CL process. 
In mobile applications, the data collected from an individual user's daily life is sparse, personalized and private, such as photos taken by user or interactions with a virtual assistant.

\noindent \textit{$\bullet$ Enhancement:}
Prior to conducting on-device CL for a new context, various enhancement techniques need to be applied to mitigate the severe impact of data scarcity, such as few-\\shot CL based on model initialization and federated CL approaches based on training algorithms. \textit{Our work focuses on the design of this stage from the data perspective.}

\noindent \textit{$\bullet$ Continual Learning}: 
The training data of both new and past contexts are mixed to update the on-device model, which has been recognized as one of the most effective methods to assimilate knowledge from new contexts without forgetting the knowledge of past contexts\footnote{It is noteworthy that our data enrichment framework can also benefit other classic CL approaches, such as parameter regularization~\cite{kirkpatrick2017overcoming, zenke2017continual} and context-adaptive model architectures~\cite{serra2018overcoming, mallya2018piggyback}, as illustrated in \S\ref{sec: Related Work}.
}~\cite{ma2023cost,buzzega2020dark,lopez2017gradient,prabhu2020gdumb, lee2022carm}.

\subsection{Limitation of Existing Approaches}
\label{sec: limitation of existing ideas}
In this section, 
we elaborate the limitations of existing few-shot CL and federated CL approaches in mitigating device-side data scarcity problem through preliminary experiments\footnote{The detailed experimental settings are introduced in \S\ref{sec: experimental setup}.}. 

\begin{figure*}
    \vspace{-0.2cm}
    \centering
    \subfigure[Performance of few-shot CL approaches without ($\blacksquare$) and with ($\square$) prior information on user contexts, and performance of vanilla CL with different amount of available training data (Vanilla-$n\times$).]{
        \includegraphics[height=3.7cm]{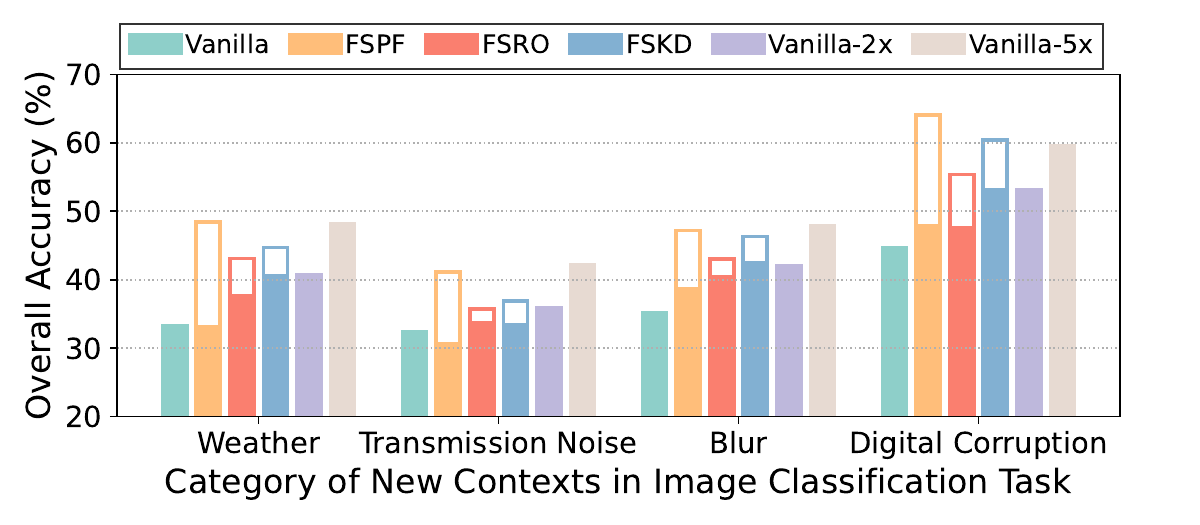}
        \label{fig: motivation_fewshotCL}
    }
    \ \ \ \ \ \ \ 
    \subfigure[Communication cost and accuracy of federated CL with varying device participation rates and data heterogeneity degree (Fed-$p$ denotes that $p\!\times\!100\%$ devices hold data from different contexts).]{
        \includegraphics[height=3.7cm]{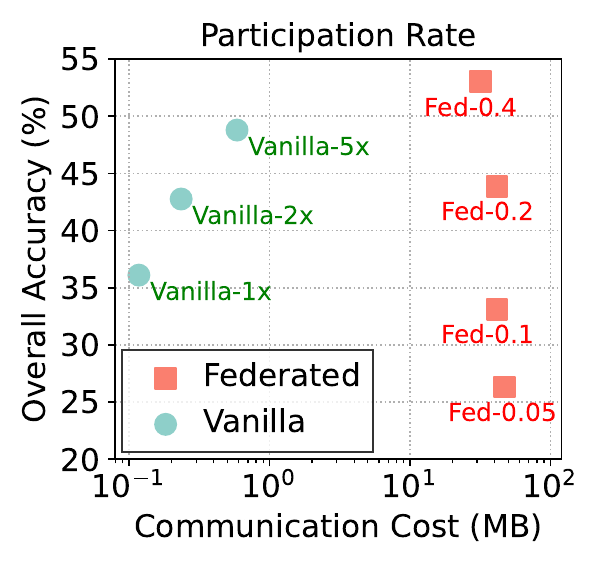}
        \includegraphics[height=3.7cm]{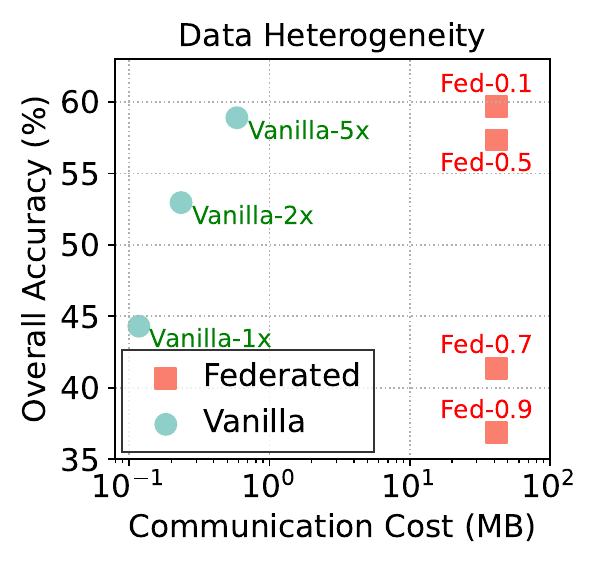}
        \label{fig: motivation_fedcl}
    }
    \vspace{-0.35cm}
    \caption{Preliminary experiments on image classification task to illustrate the limitations of existing solutions.}
    \Description{Preliminary experiments on image classification task to illustrate the limitations of existing solutions.}
    \vspace{-0.34cm}
    \label{fig: limitation of related work}
\end{figure*}

\textbf{Few-shot CL}~\cite{mazumder2021few, shi2021overcoming, zhao2023few} proposes pretraining ML models on base contexts with massive public data to capture general knowledge, which is then transferred to new contexts through transfer learning techniques. 
Representative methods include:
    1) knowledge distillation~\cite{zhao2023few} ({\ttfamily FS-KD}), which distills past contexts' knowledge to the new context's model by keeping the model outputs of historical data samples unchanged, 
    2) robust optimization~\cite{shi2021overcoming} ({\ttfamily FS-RO}), which constrains model parameters within the common flat minima of all contexts' training objective functions, and 
    3) parameter freezing~\cite{mazumder2021few} ({\ttfamily FS-PF}), which freezes the important parameters with high value of the previously trained model.

However, most of these few-shot CL approaches depend on a powerful model pre-training process, which pretrain either a large model on data from diverse contexts to fully capture the general knowledge, or a tiny model on a customized dataset to learn personalized knowledge. Unfortunately, both of them are impractical for on-device scenarios due to limited hardware resources and unpredictable contexts.
On one hand, the limited memory and computational capabilities of mobile devices restrict the size and capacity of deployed models, impeding effective model pretraining over diverse data.  
On the other hand, the uncertainty of future user contexts prevents the pre-selection of a tailored data-subset for pretraining before model deployment.
Our preliminary experiments shown in Figure \ref{fig: motivation_fewshotCL} reveal that the performance of few-shot CL declines significantly without prior information on user contexts, with model accuracy reduction ranging from $8.6\!-\!15.3\%$ for {\ttfamily FS-PF}, $1.9\!-\!7.9\%$ for {\ttfamily FS-RO} and $3.9\!-\!7.2\%$ for {\ttfamily FS-KD}. 
In contrast, simply increasing the training data size to $50$ can outperform all few-shot CL approaches, underscoring the potential of data enrichment.

\textbf{Federated CL}~\cite{yoon2021federated, dong2022federated} utilizes a cloud server to periodically aggregate the parameters of models trained on distributed devices, thereby mitigating the overfitting issue on individual devices and facilitating knowledge transfer across multiple devices.
However, the substantial communication overheads and unstable model training process render federated CL impractical for mobile devices.
First, the frequent exchange of model parameters between mobile devices and the cloud server incurs significant communication costs and prolongs the wall-clock training time for on-device models.
Second, the model performance of federated CL is relatively sensitive to the device participation rate (or amount) and the data heterogeneity across devices~\cite{li2022pyramidfl, shin2022fedbalancer}.
Experimental results shown in Figure \ref{fig: motivation_fedcl} indicate that: 
1) Federated CL achieves superior performance only when $\ge\!20\%$ devices participate in each round of model aggregation or when more than $\ge\!30\%$ mobile users experience similar contexts, which can be unrealistic in real-world settings;
2) In comparison to federated CL, transmitting data with a suitable distribution from cloud to each device could reach the same target accuracy with communication costs reduced to less than $1\%$.

\section{Problem Definition}
In this section, we present a generic formalization of the cloud-assisted data enrichment problem for on-device CL. We consider a scenario where a mobile user sequentially encounters $T$ new contexts. Each context $t\!=\!1,\dots,T$ has an underlying data distribution $\mathcal{D}_{de}^t$ and the device collects an empirical dataset $\widehat{\mathcal{D}}_{de}^t$ for training on-device model.
Due to the scarcity of user data, a similar data-subset $\mathcal{S}^t$ is expected to be retrieved from the cloud-side dataset $\mathcal{D}_{cl}$ to enrich the on-device empirical dataset $\widehat{\mathcal{D}}_{de}^t$ and thereby enhance the CL performance. 

To assess the effectiveness of data enrichment, we first define a metric to evaluate the similarity between two datasets in terms of their impacts on the model training process.
In on-device CL, model parameters are typically fine-tuned by on-device data via gradient descent methods. Therefore, the similarity between two datasets $\mathcal{D}_1$ and $\mathcal{D}_2$ with respect to the training process of model $\theta$ can be quantified by the maximal difference between the average gradients of $\mathcal{D}_1$ and $\mathcal{D}_2$ within a nearby parameter space $\{\theta'|\!\parallel\!\theta'\!-\!\theta\!\parallel\le\epsilon\}$:
\begin{equation}
    \begin{small}
        \begin{aligned}
            Sim(\mathcal{D}_1,\mathcal{D}_2|\theta)\triangleq-\max_{\parallel\theta'-\theta\parallel\le\epsilon}\big|\big|\nabla L(\mathcal{D}_1,\theta')-\nabla L(\mathcal{D}_2,\theta')\big|\big|,
        \end{aligned}
    \end{small}
    \label{eq: dataset similarity}
    \vspace{-0.1cm}
\end{equation}
where $L(\mathcal{D},\theta)\!=\!\mathbb{E}_{(x,y)\in\mathcal{D}}\big[l(x,y,\theta)\big]$ denotes the expected loss of model $\theta$ over dataset $\mathcal{D}$. A high similarity between two datasets $\mathcal{D}_1$ and $\mathcal{D}_2$ implies their comparable performance in updating model parameters for multiple steps, resulting in similar impacts on on-device model training.

\textbf{Problem Formulation.}
For each new context $t$, the cloud server aims to select the most similar data-subset $\mathcal{S}^{t,*}\!\subseteq\!\mathcal{D}_{cl}$ to update the current on-device model $\theta^{t-1}$ in a similar way with the device-side underlying data distribution $\mathcal{D}_{de}^t$, which means that $\mathcal{S}^{t,*}$ and $\mathcal{D}_{de}^t$ should exhibit high similarity as measured by the metric in Equation (\ref{eq: dataset similarity}). Consequently, the data enrichment problem can be formally expressed as:
\begin{equation}\label{eq: data enrichment problem}
    \begin{small}
    \begin{aligned}
        \mathcal{S}^{t,*}= & \mathop{\arg\max}\limits_{\mathcal{S}^t\subseteq \mathcal{D}_{cl},|\mathcal{S}^t|\le B}  Sim(\mathcal{S}^t,\mathcal{D}_{de}^t\ |\ \theta^{t-1})\\
        \approx & \mathop{\arg\max}\limits_{\mathcal{S}^t\subseteq\mathcal{D}_{cl},|\mathcal{S}^t|\le B} Sim(\mathcal{S}^t,\widehat{\mathcal{D}}_{de}^t\ |\ \theta^{t-1}),
    \end{aligned}
    \end{small}
\end{equation}
where $B$ represents the maximum allowable size of the selected data-subset and is constrained by the communication cost budget of each device.
This formulation enables the device to enhance model training performance by expanding the training data from the collected dataset $\widehat{\mathcal{D}}_{de}^t$ to the enriched larger-scale dataset $\mathcal{S}^t$, while ensuring that the enriched data follows a similar distribution. 

\textbf{Practical Challenges.}
Directly solving the data enrichment problem in Equation (\ref{eq: data enrichment problem}) brings severe privacy concerns for mobile users and high computational burden for cloud server. 
First, the mobile device needs to upload both the current model $\theta^{t-1}$ and raw user data $\widehat{\mathcal{D}}_{de}^t$ to the cloud server, which poses a severe breach of user privacy. Second, the cloud server has to compute the similarity score $Sim(\mathcal{S}^t,\widehat{\mathcal{D}}_{de}^t|\theta^{t-1})$ for every possible data-subset $\mathcal{S}^t\!\subseteq\!\mathcal{D}_{cl}$, $|\mathcal{S}^t|\!\le\!B$, resulting in exponential computation complexity. 
\section{Framework Design}
\label{sec: framework design}
\begin{figure}
    \centering
    \vspace{-0.15cm}
    \includegraphics[width=0.45\textwidth]{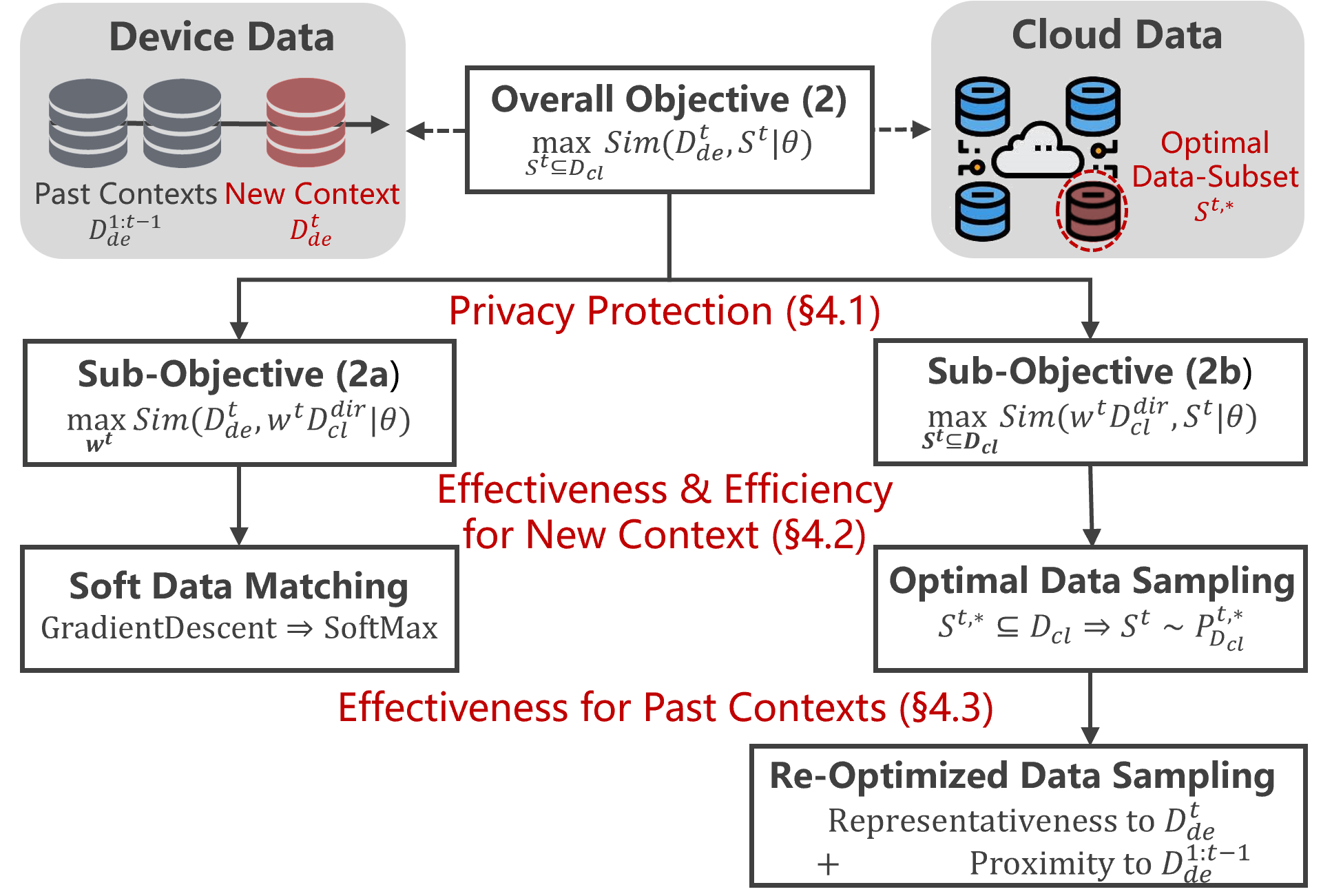}
    \vspace{-0.3cm}
    \caption{Design Rationale of {\ttfamily Delta}.}
    \Description{Design Rationale of {\ttfamily Delta}.}
    \vspace{-0.4cm}
    \label{fig: designrationale}
\end{figure}

{\ttfamily Delta} incorporates three key components to render data enrichment systematically practical: the construction of a directory dataset to address privacy concerns (\S\ref{sec: cloud-side directory dataset}), device-side soft data matching strategy coupled with a cloud-side data sampling scheme to efficiently and effectively enrich data for new contexts (\S\ref{sec: data enrichment for new scenario}), and a re-optimization of the cloud-side data sampling to further enhance its effectiveness across both past and new contexts (\S\ref{sec: data enrichment for past scenarios}). \textit{Each component is inspired and supported by theoretical analysis presented in \S\ref{sec: theoretical analysis} and the overall design rationale is illustrated in Figure \ref{fig: designrationale}.}

\subsection{Directory Dataset Construction}
\label{sec: cloud-side directory dataset}
To address privacy concerns, {\ttfamily Delta} introduces the concept of ``directory'' dataset, which facilitates decomposing the data enrichment problem (\ref{eq: data enrichment problem}) into two sub-problems, and allows the device and cloud to collaboratively solve the sub-problems without the need to share raw user data.

\textbf{Design Rationale.}
Inspired by the directory structures in storage systems~\cite{daley1965general}, {\ttfamily Delta} constructs a compact directory dataset consisting of a few data samples to represent the extensive cloud-side dataset, denoted as $\mathcal{D}_{cl}^{dir}\!=\!\big\{(\bar{x}_c,\bar{y}_c)\big\}_{c=1}^{|\mathcal{D}_{cl}^{dir}|}$. This directory dataset can be pre-downloaded by mobile devices along with the model deployment.
As illustrated in Figure \ref{fig: designrationale} and supported in Theorem \ref{theorem: problem decomposition}, the objective function (\ref{eq: data enrichment problem}) of the data enrichment problem can be decomposed into the sum of two sub-objective functions:
\begin{itemize}[leftmargin=0.3cm, topsep=0cm]
    \item \textit{Sub-objective (2a)}: similarity between the device-side dataset $\mathcal{D}_{de}^t$ and the weighted directory dataset $w^t\mathcal{D}_{cl}^{dir}$, where each data sample $(\bar{x}_c,\bar{y}_c)\!\in\!\mathcal{D}_{cl}^{dir}$ is assigned a weight $w_c^t$. The weight vector $w^t$ is a variable to be optimized. 
    \item \textit{Sub-objective (2b)}: similarity between the weighted directory dataset $w^t\mathcal{D}_{cl}^{dir}$ and the cloud-side data-subset $\mathcal{S}^t$, where $\mathcal{S}^t$ is the variable to be optimized. 
\end{itemize}
\noindent These two two sub-objective functions can be optimized sequentially and independently by the mobile device and cloud server through the exchange of non-sensitive information:

\noindent \textit{1) Mobile device} optimizes sub-objective (2a) by computing the optimal weight $w^{t,*}$ for the directory dataset $\mathcal{D}_{cl}^{dir}$ to represent the device-side data distribution $\mathcal{D}_{de}^t$. 

\noindent \textit{2) Cloud server} optimizes sub-objective (2b) by searching for the optimal cloud-side data-subset $\mathcal{S}^{t,*}\!\subseteq\!\mathcal{D}_{cl}$ to align with the weighted directory dataset $w^{t,*}\mathcal{D}_{cl}^{dir}$, with $w^{t,*}$ being uploaded by the mobile device after device-side optimization.

\noindent \textit{3) Device-cloud communication} involves the cloud-side directory dataset $\mathcal{D}_{cl}^{dir}$ and the device-side optimized weight $w^{t,*}$, which do not involve any raw user data and thus protect user privacy akin to classic federated learning~\cite{mcmahan2017communication}. Detailed discussion and comparison are presented in \S\ref{sec: limitation and future work}.

\textbf{Practical Implementation.}
The practical effectiveness of the above decomposition process relies on an appropriate directory dataset that accurately represents the cloud-side public dataset. 
While classical data clustering methods can be used to select cluster centroids as the directory dataset elements, we observe that directly clustering raw data samples may not fully capture the influence of data on model training, due to the diverse sources, wide-ranging distributions and varying dimensions of cloud-side data.
To address this issue, we take advantage of the typical paradigm of on-device model training~\cite{cai2023efficient, cai2020tinytl,dettmers2024qlora}, where the feature extractor $\phi$ is pre-trained on extensive cloud-side data for generalization ability and the classifier $\psi$ is trained on device-side data for personalization performance. 
We propose clustering data samples $(x,y)\!\in\!\mathcal{D}_{cl}$ based on the feature extractor outputs $\phi(x)$ rather than raw input $x$, and selecting the cluster centroids as elements of the directory dataset, which offers two advantages: 
1) features as model's intermediate outputs have a consistent dimension and are more relevant to model training than raw inputs, 
2) the features of most cloud-side data samples are already available from the pre-training process of feature extractor, incurring minimal additional costs.
\begin{figure*}
    \vspace{-0.2cm}
    \centering
    \includegraphics[width=0.85\textwidth]{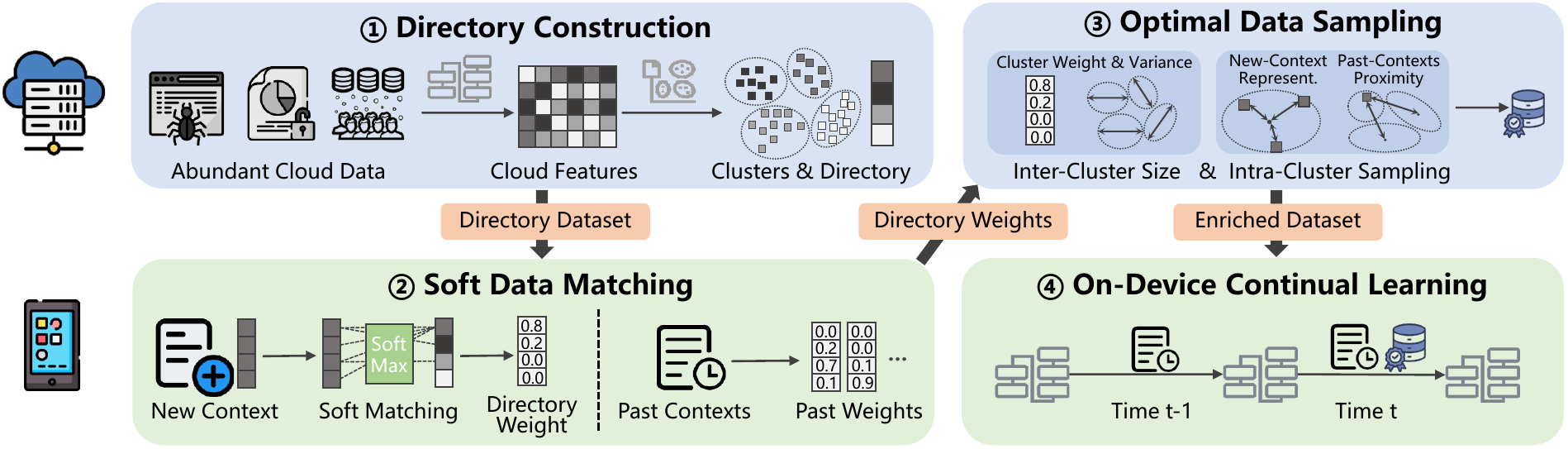}
    \vspace{-0.2cm}
    \caption{Overall Workflow of {\ttfamily Delta} Framework. {\ttfamily Delta} serves as a plug-in for on-device continual learning.}
    \Description{Overall Workflow of {\ttfamily Delta} Framework. {\ttfamily Delta} serves as a plug-in for on-device continual learning.}
    \vspace{-0.35cm}
    \label{fig: delta framework}
\end{figure*}

\subsection{Data Enrichment for New Context}
\label{sec: data enrichment for new scenario}
While the directory dataset safeguards user privacy by decomposing the data enrichment problem into device-side and cloud-side sub-problems, it is non-trivial to solve them in an effective and efficient manner due to the scarcity of on-device data and diversity of cloud-side data.
\begin{itemize}[leftmargin=0.3cm, topsep=0.cm]
    \item \textit{Device-side ineffectiveness}: Solving sub-problem (2a) requires determining the optimal weight $w^{t,*}$ to align the weighted directory dataset $w^t\mathcal{D}_{cl}^{dir}$ with the device-side data distribution $\mathcal{D}_{de}^t$. However, the underlying data distribution is typically approximated by the sparse empirical dataset $\widehat{\mathcal{D}}_{de}^t$ stored by mobile device, which can cause conventional gradient descent algorithms to converge to local optima. Consequently, the derived weight becomes overfitted to the limited empirical dataset and ineffective in representing the device-side data distribution.
    \item \textit{Cloud-side inefficiency}: Exactly solving sub-problem (2b) involves evaluating the similarity score for each potential cloud-side data subset, which requires exploring a vast feasible region of candidate data-subsets $\mathcal{S}^{t}\!\subseteq\!\mathcal{D}_{cl},|\mathcal{S}|\!\le\!B$ and results in exponential computation and time complexity for cloud server, leading to low efficiency.
\end{itemize}
\noindent To achieve an efficient and effective data enrichment process for each coming context, we propose a soft data matching strategy for mobile device to derive a representative directory weight by fully leveraging the limited on-device data, and a data sampling scheme for cloud server to sample an optimal data-subset with constant time complexity.

\textbf{Device-Side: Soft Data Matching.} 
To prevent the directory weight $w^t$ from overfitting to scarce on-device data, 
we propose to assign physical meanings to $w^t$ by interpreting each element $w^t_c$ as the fraction of on-device data that exhibits high similarity with the cloud-side cluster centroid $(\bar{x}_c,\bar{y}_c)\!\in\!\mathcal{D}_{cl}^{dir}$. Thus, for each data sample $(x,y)\!\in\!\widehat{\mathcal{D}}_{de}^t$ collected by mobile device, its similarities with all the cluster centroids are computed, and the weight of the most similar one is incremented by one step:
\begin{equation}
    \begin{small}
        \begin{aligned}
        &  
        c^*= \mathop{\arg\max}_{c}Sim\big((x,y), (\bar{x}_c,\bar{y}_c)\ |\ \theta^{t-1}\big),\\
        & w_{c^*}^t\gets w_{c^*}^t+1. & \mathrm{(Hard\ Matching)}
        \end{aligned}
    \end{small}
    \nonumber
\end{equation}

\noindent However, in our experiments, we observe that each on-device data sample can exhibit high similarity with more than one cloud-side cluster centroids, which is influenced by the granularity of cloud-side data clustering (\textit{i.e.} the number of data clusters) during the directory construction process.
However, the ``hard'' matching function $argmax$ is incapable of capturing the correlation between one device-side sample and multiple cloud-side clusters. Thus, we propose to employ a ``soft'' matching function $softmax$, allowing each data sample to contribute to the weights of more than one clusters:
\begin{equation}
    \begin{small}
        \begin{aligned}
            \forall c, w_c^t\gets w_c^t+Softmax\left(\frac{Sim\big((x,y),(\bar{x}_c,\bar{y}_c)\ |\ \theta^{t-1}\big)}{\tau}\right),        
        \end{aligned}
    \end{small}
    \label{eq: soft matching}
\end{equation}
where $\tau$ is a temperature hyperparameter to control the weight increments of clusters with different degrees of similarity. As $\tau\rightarrow 0$, $softmax$ gradually degrades to $argmax$.

\textbf{Cloud-Side: Optimal Data Sampling.}
To enhance efficiency and reduce the computational overhead on the cloud server, we propose transforming the ``hard'' data selection process into a ``soft'' data sampling process. The key difference is that the former seeks to find an exact data-subset $\mathcal{S}^{t,*}$ to optimize sub-problem (2b), whereas the latter aims to compute a data sampling policy $P_{\mathcal{D}_{cl}}^{t,*}$ such that the sampled data-subset is optimal for sub-problem (2b) in expectation:
\begin{equation}
    \begin{small}
    \begin{aligned}
        \mathop{\max}_{\mathcal{S}^t\subseteq\mathcal{D}_{cl}} & Sim(\mathcal{S}^t,w^{t}\mathcal{D}_{cl}^{dir}|\theta^{t-1}) & \mathrm{(Hard\ Selection)}\\
        \Rightarrow
        \mathop{\max}_{P_{\mathcal{D}_{cl}}^t} \ \ & \mathbb{E}_{\mathcal{S}^t\sim P_{\mathcal{D}_{cl}}^t}\big[Sim(\mathcal{S}^t,w^{t}\mathcal{D}_{cl}^{dir}|\theta^{t-1})\big]. & \mathrm{(Soft\ Sampling)}
    \end{aligned}
    \end{small}
    \nonumber
\end{equation}
This transformation allows the cloud server to directly identify an appropriate data-subset through data sampling policy, which can be computed with constant time complexity.

We outline the specific operations of cloud-side data sampling scheme, with theoretical foundation provided in \S\ref{sec: theoretical analysis for new context}. 
The scheme involves \textit{inter-cluster size allocation} and \textit{intra-cluster data sampling}, which determine \textit{how many} and \textit{which} data samples to select from each cloud-side data cluster:

    \noindent \textit{$\bullet$ Inter-cluster size allocation.}
    Given that the size of the selected data-subset is limited by the communication cost budget, the cloud server needs to allocate distinct sampling sizes to different data clusters to maximize the overall similarity between the sampled data-subset and the weighted directory dataset, \textit{i.e.} sub-objective (2b). 
    As demonstrated in Lemma \ref{lemma: optimal data sampling}, the optimal sampling size $|\mathcal{S}^{t,*}_c|$ for each cluster $\mathcal{D}_{cl,c}$ depends on its directory weight $w_c^t$ and the dispersion degree of intra-cluster feature distribution $\mathbb{E}_x||\phi(x)\!-\!\phi(\bar{x})||$:
    \begin{equation}
        \begin{small}
        \begin{aligned}
            |\mathcal{S}^{t,*}_c|\propto w_c^t\cdot\mathbb{E}_{(x,y)\in\mathcal{D}_{cl,c}}\big|\big|\phi(x)-\phi(\bar{x}_c)\big|\big|.
        \end{aligned}
        \end{small}
        \label{eq: inter-cluster size allocation}
    \end{equation}
    For each cluster, a higher weight suggests a higher similarity with the device-side data for on-device CL, and a wider feature distribution indicates the need for more data samples to comprehensively represent the cluster .
    
    \noindent \textit{$\bullet$ Intra-Cluster Data Sampling.}
    Within each cloud-side data cluster $\mathcal{D}_{cl,c}$, the optimal sampling probability for each data sample $(x,y)$ is proportional to the feature distance between such data sample and the cluster centroid $(\bar{x}_c,\bar{y}_c)$:
    \begin{equation}
        \begin{small}
            \begin{aligned}
                P_{\mathcal{D}_{cl,c}}^{t,*}(x,y)=\frac{\big|\big|\phi(x)-\phi(\bar{x}_c)\big|\big|}{\sum_{(x',y')\in\mathcal{D}_{cl,c}}\big|\big|\phi(x')-\phi(\bar{x}_c)\big|\big|}.
            \end{aligned}
        \end{small}
    \label{eq: intra-cluster data sampling}
    \end{equation}
    Theoretically, our analysis in Lemma \ref{lemma: optimal data sampling} demonstrates that this sampling probability could maximize the expected similarity between each data cluster $\mathcal{D}_{cl,c}$ and the corresponding selected data-subset $\mathcal{S}^t_c$, thereby optimizing sub-objective (2b) in expectation given fixed directory weights $w^t$.
    Intuitively, this sampling strategy favors data samples that are farther from the cluster centroid, which enhances the diversity and informativeness of the selected data-subset while ensuring unbiasedness and representativeness through data re-weighting technique like importance sampling~\cite{katharopoulos2018not}.

\subsection{Data Enrichment for All Contexts}
\label{sec: data enrichment for past scenarios}
Although the previous components ensure a private, efficient and effective data enrichment process for each new context, the notorious issue of catastrophic forgetting (\textit{i.e.} inferior memory stability) is also exacerbated.
First, as model parameters $\theta$ continually adapt to the enriched data $\{\mathcal{S}^i\}_{i=1}^{t}$, the similarity between each past context $i$'s enriched data $\mathcal{S}^i$ and the underlying distribution $\mathcal{D}_{de}^i$ gradually diminishes, hindering the use of $\{\mathcal{S}^i\}_{i=1}^{t}$ for retaining past knowledge.
Second, independently enriching data solely for the new context will exacerbate the mutual interference between the model training processes of new and past contexts.

To address these issues, we take the first step to theoretically analyze the correlation between new context's enriched data and the model performance on both new and past contexts. 
Further, we re-optimize the data sampling scheme for cloud server to identify a data-subset that could contribute to the learning processes of both new and past contexts.

\textbf{Theoretical Analysis.} Theorem \ref{theorem: relationship between past and new scenarios} reveals that the overall CL performance, quantified by the average loss of model over all contexts, is primarily determined by three terms:

\noindent \textit{1) New-context representativeness}, which is quantified by the feature distance between the enriched dataset $\mathcal{S}^t$ and the underlying data distribution of new context $\mathcal{D}_{de}^t$.

\noindent \textit{2) Past-contexts proximity}, which is measured by the feature distance between the enriched dataset $\mathcal{S}^t$ and the underlying data distributions of all the past contexts $\{\mathcal{D}_{de}^i\}_{i=1}^{t-1}$.

\noindent \textit{3) Cross-Context Heterogeneity}, which is a fixed term and determined by the heterogeneity between the new context and the past contexts encountered by the mobile user.

\noindent Consequently, the original intra-cluster data sampling strategy in Equation (\ref{eq: intra-cluster data sampling}) can be seen as focusing only on the first term (\textit{i.e.} effectiveness for new context), while overlooking the second term (\textit{i.e.} effectiveness for past contexts.)


\textbf{Practical Implementation.}
Guided by the theoretical results, we further derive the analytical expression for the re-optimized cloud-side data sampling policy, with the detailed mathematical derivation provided in \S\ref{sec: theoretical analysis for past contexts}. Specifically, for intra-cluster data sampling, the optimal sampling probability for each data sample is proportional to the weighted sum of new-context representativeness and past-contexts proximity:
\begin{equation*}
    \begin{small}
    \begin{aligned}
        & P_{\mathcal{D}_{cl,c}}^{t,*}(x,y)
        \propto \Big|\Big|\phi(x)-\phi(\bar{x}_c)\Big|\Big|+\alpha\Bigg|\Bigg|\phi(x)-\frac{\sum_{i=1}^{t-1}\phi(w^{i,*}\mathcal{D}_{cloud}^{dir})}{t-1}\Bigg|\Bigg|,
    \end{aligned}
    \end{small}
    \label{eq: re-optimized intra-cluster data sampling}
\end{equation*}
where $\alpha$ is a hyperparameter determined by the device to balance the model performance over the new context and past contexts when conducing cloud-assisted data enrichment.

\subsection{Overall Framework}
\label{sec: overall framework}

We illustrate the overall workflow of {\ttfamily Delta} framework  in Figure \ref{fig: delta framework}, which comprises four stages.
\noindent\textbf{\ding{202} Directory Construction:}
Initially, the cloud server utilizes the pre-trained feature extractor to extract features from diverse datasets and performs data clustering to construct the directory dataset. The directory dataset is then distributed to mobile devices along with the model deployment. 
\noindent\textbf{\ding{203} Soft Data Matching:}
For each coming new context $t$, the mobile device solves sub-problem (2a) through the soft data matching strategy outlined in Equation (\ref{eq: soft matching}), and uploads the optimal directory weights for both the new and past contexts to the cloud server.
\noindent\textbf{\ding{204} Optimal Data Sampling:}
Upon receiving the directory weights, the cloud server computes the analytical expressions for the optimal data sampling scheme, which includes inter-cluster size allocation in Equation (\ref{eq: inter-cluster size allocation}) and intra-cluster data sampling in Equation (\ref{eq: intra-cluster data sampling}). The optimal data-subset is then sampled according to the scheme and transmitted back to the mobile device.
\noindent\textbf{\ding{205} On-Device Continual Learning:}
The mobile device conducts CL process using the enriched datasets of both new and past contexts. 
\textit{Generally, {\ttfamily Delta} serves as a plug-in module to enhance on-device CL performance with privacy protection, effectiveness and efficiency.}
\section{Theoretical Analysis}
\label{sec: theoretical analysis}
In this section, we provide theoretical foundations for the key components of {\ttfamily Delta} framework. Detailed proofs are provided in supplementary material~\cite{technical_report} due to limited space.

\subsection{Theory for Directory Construction}
\label{sec: theoretical analysis for directory construction}
To facilitate data enrichment as outlined in Equation (\ref{eq: data enrichment problem}) without disclosing raw user data, we introduce the directory dataset to decompose the original objective function into two sub-objective functions. The performance of this decomposition is theoretically guaranteed by Theorem \ref{theorem: problem decomposition}, which elucidates the relation between the original objective function and two sub-objective functions.
$\vspace{-0.15cm}$
\begin{theorem}\label{theorem: problem decomposition}
    Given directory dataset $\mathcal{D}_{cl}^{dir}$, the maximal similarity between the device-side dataset $\mathcal{D}_{de}^t$ and the cloud-side data-subset $\mathcal{S}^{t}\!\subseteq\!\mathcal{D}_{cl}$ for model $\theta^{t-1}$ can be bounded by
    \begin{equation}\label{eq: decomposition}
        \begin{small}
        \begin{aligned}
            \underbrace{\max_{\mathcal{S}^t\subseteq\mathcal{D}_{cl}}Sim(\mathcal{D}_{de}^t,\mathcal{S}^{t}|\theta^{t-1})}_{\mathrm{original\ objective}}
            \ge  & \underbrace{\max_{w^t}Sim(\mathcal{D}_{de}^t,w^t\mathcal{D}_{cl}^{dir}|\theta^{t-1})}_{\mathrm{sub-objective\ (2a)\ for\ optimal\ weight}} \\
            & +\underbrace{\max_{\mathcal{S}^t\subseteq\mathcal{D}_{cl}}Sim(\mathcal{S}^t,w^{t,*}\mathcal{D}_{cl}^{dir}|\theta^{t-1})}_{\mathrm{sub-objective\ (2b)\ for\ optimal\ subset}},
        \end{aligned}
        \end{small}
        \nonumber
        \vspace{-0.1cm}
    \end{equation}
    where $w^t\mathcal{D}_{cl}^{dir}$ represents the weighted directory dataset.
    $\vspace{-0.15cm}$
\end{theorem}
\noindent This theorem shows that the optimal value of the original objective function (\ref{eq: data enrichment problem}) is bounded from below by the sum of the optimal values of the two sub-objective functions (2a) and (2b). Consequently, {\ttfamily Delta} essentially optimizes the worst-case performance of data enrichment for diverse contexts. 
The practical gap between the original and decomposed objective functions is determined by the representativeness of the cloud-side directory dataset. 
In \S\ref{sec: component-wise analysis}, we empirically show that a directory dataset with around $10^2$ elements is sufficient to represent a cloud-side dataset consisting of $10^6$ data samples across $10^2$ contexts. 

\subsection{Theory for New Context's Enrichment}
\label{sec: theoretical analysis for new context}
To provide theoretical guarantees for the optimality of the cloud-side data sampling scheme outlined in Equations (\ref{eq: inter-cluster size allocation}) and (\ref{eq: intra-cluster data sampling}), we first present Theorem \ref{theorem: cloud-side data selection} to partition the on-device model $\theta$ into a feature extractor $\phi$ and a classifier $\psi$ with a Lipstchiz continuity constant $L_\psi$. The feature extractor is typically pre-trained by cloud server and remains unchanged during the on-device model training process.
$\vspace{-0.15cm}$
\begin{theorem}\label{theorem: cloud-side data selection} 
    The expected similarity between the weighted directory dataset $w^t\mathcal{D}_{cl}^{dir}$ and the data-subset $\mathcal{S}^t$ selected according to sampling scheme $P_{\mathcal{D}_{cl}}^t$ is bounded by:
    $\vspace{-0.1cm}$
    \begin{equation}
        \begin{small}
        \begin{aligned}
            & \mathbb{E}_{\mathcal{S}^t\sim P_{\mathcal{D}_{cl}}^t}\big[Sim(\mathcal{S}^t,w^t\mathcal{D}_{cl}^{dir}\ |\ \theta^{t-1})\big]\\
            \ge & -\mathbb{E}_{\mathcal{S}^t\sim P_{\mathcal{D}_{cl}}^t}L_{\psi}\Big|\Big|\mathbb{E}_{(x,y)\in\mathcal{S}^t}\big[\phi(x)\big]-\sum_cw_c^t\phi(\bar{x}_c)\Big|\Big|.
        \end{aligned}
        \end{small}
        \nonumber
        \vspace{-0.1cm}
    \end{equation}
\end{theorem}
\noindent Further, in Lemma \ref{lemma: optimal data sampling}, we demonstrate that the expected value of sub-objective function (2b) (i.e., the lower bound
of the above inequality) is determined by two terms: inter-cluster sampling size $|\mathcal{S}^t_c|$ and intra-cluster sampling probability $P_{\mathcal{D}_{cl,c}}^t(x,y)$ for each cloud-side data cluster $c$. 
$\vspace{-0.15cm}$
\begin{lemma}
    \label{lemma: optimal data sampling}
    The expected similarity between the sampled data-subset $\mathcal{S}^t$ and the weighted directory dataset $w^t\mathcal{D}_{cl}^{dir}$ is determined by each cluster $c$'s sampling size $|\mathcal{S}_c^t|$ and intra-cluster data sampling probability $P_{\mathcal{D}_{cl,c}}^t(x,y)$:
    $\vspace{-0.1cm}$ 
    \begin{equation}
        \begin{small}
        \begin{aligned}
            & \mathop{\min}\limits_{P_{\mathcal{D}_{cl}}^t}   \mathbb{E}_{\mathcal{S}^t\sim P_{\mathcal{D}_{cl}}^t}\Big|\Big|\mathbb{E}_{(x,y)\in\mathcal{S}^t}\big[\phi(x)\big]-\sum_cw_c^t\phi(\bar{x}_c)\Big|\Big|\\
            = & \mathop{\min}\limits_{|\mathcal{S}^t_c|, P_{\mathcal{D}_{cl,c}}^t}\sum_c\bigg(\frac{(w_c^t)^2}{|\mathcal{S}_c^t|}\cdot\sum_{(x,y)\in \mathcal{D}_{cl,c}}\frac{\big|\big|\phi(x)-\phi(\bar{x})\big|\big|^2}{|\mathcal{D}_{cl,c}|^2\cdot P_{\mathcal{D}_{cl,c}}^t(x,y)}\bigg)
        \end{aligned}
        \end{small}
        \nonumber
        \vspace{-0.1cm}
    \end{equation}
\end{lemma}
\noindent By leveraging Cauchy-Schwarz inequality, we can derive the analytical expressions of the optimal data sampling policy (i.e. $|\mathcal{S}^{t,*}_c|$ and $P_{\mathcal{D}_{cl}}^{t,*}$), which can be computed directly using the directory weights uploaded by mobile device:
\begin{equation}
    \begin{small}
    \begin{aligned}
        \begin{cases}
                \ \ \ \quad|\mathcal{S}_c^{t,*}|\ \ \quad\propto\ \ w_c^t\cdot\mathbb{E}_{(x,y)\in\mathcal{D}_{cl,c}}\big|\big|\phi(x)-\phi(\bar{x}_c)\big|\big|\\
                \ \mathcal{P}_{\mathcal{D}_{cl,c}}^{t,*}(x,y)\ \ \propto\ \ \big|\big|\phi(x)-\phi(\bar{x}_c)\big|\big|,\ \ \forall (x,y)\in\mathcal{D}_{cl,c}.
            \end{cases}
    \end{aligned}
    \end{small}
\end{equation}


\subsection{Theory for All Contexts' Enrichment}
\label{sec: theoretical analysis for past contexts}
In \S\ref{sec: data enrichment for past scenarios}, we propose to refine the cloud-side data sampling scheme to ensure that the enriched data for new context can contribute to the learning processes of both new and past contexts. 
To achieve this, we first analyze the impact of new context's enriched data on the model performance over all contexts in Theorem \ref{theorem: relationship between past and new scenarios}, which consists of three key terms: representativeness to new context, proximity to past contexts and the data heterogeneity across different contexts.
$\vspace{-0.15cm}$
\begin{theorem}\label{theorem: relationship between past and new scenarios}
    In $m$-th training round for context $t$, when the model parameters are updated from $\theta^{t,m}$ to $\theta^{t,m+1}$ using the enriched data $S^t$ sampled by policy $P_{\mathcal{D}_{cl}}^t$, the expected reduction in model loss (or improvement in model performance) over all contexts' data distribution $\mathcal{D}_{de}^{1:t}$ can be bounded by:
    $\vspace{-0.1cm}$
    \begin{equation}
        \begin{footnotesize}
        \begin{aligned}
            & \mathbb{E}_{\mathcal{S}^t\sim P_{\mathcal{D}_{cl}}^t}\Big[\underbrace{L(\mathcal{D}_{de}^{1:t},\theta^{t,m+1})-L(\mathcal{D}_{de}^{1:t},\theta^{t,m})}_{\mathrm{loss\ reduction\ in\ }m\mathrm{-th\ model\ update}}\Big]\\
            \le & \frac{1}{2}(H\eta^2-\eta)L_{\psi}\underbrace{\mathbb{V}_{\mathcal{S}^t\sim P_{\mathcal{D}_{cl}}^t}\big[\phi(\mathcal{D}_{de}^t)-\phi(\mathcal{S}^t)\big]}_{\mathrm{representativeness\ to\ new\ context\ }t}+\\
            & \frac{\eta L_{\psi}}{2}\underbrace{\mathbb{V}_{\mathcal{S}^t\sim P_{\mathcal{D}_{cl}}^t}\big[\phi(\mathcal{D}_{de}^{1:t-1})\!-\!\phi(\mathcal{S}^t)\big]}_{\mathrm{proximity\ to\ past\ contexts\ }1\sim t-1}
            +\frac{\eta L_{\psi}}{2}\underbrace{\big|\big|\phi(\mathcal{D}_{de}^t)\!-\!\phi(\mathcal{D}_{de}^{1:t-1})\big|\big|^2}_\mathrm{heterogeneity\ across\ contexts},
        \end{aligned}
        \end{footnotesize}
        \nonumber
    \end{equation}
    where $\mathbb{V}_{x}[f(x)]$ denotes the variance of function $f(x)$.
    \vspace{-0.12cm}
\end{theorem}
\noindent Building on this analysis, we observe that to improve the overall CL performance and reduce the model loss across all contexts, the cloud-side sampling scheme $P_{\mathcal{D}_{cl}}^t$ should take both the representatievess to new context and the  proximity to past contexts into consideration. From a theoretical perspective, we further derive the analytical expression of the re-optimized data sampling scheme $P_{\mathcal{D}_{cl}}^{t,*}$ in Lemma \ref{lemma: re-optimal cloud-side data selection}.
$\vspace{-0.15cm}$
\begin{lemma}\label{lemma: re-optimal cloud-side data selection}
    To optimize the model performance on the overall data distribution of all encountered contexts, the intra-cluster data sampling probability $P_{\mathcal{D}_{cl}}^{t,*}$ needs to be refined as:
    \begin{equation}
        \begin{small}
        \begin{aligned}
            P_{\mathcal{D}_{cl}}^{t,*}(x,y)\propto \sqrt{\big|\big|\phi(x)-\phi(\bar{x}_c)\big|\big|^2\!+\!\alpha\big|\big|\phi(x)-\phi(\mathcal{D}_{de}^{1:t-1})\big|\big|^2},
        \end{aligned}
        \end{small}
        \nonumber
    \end{equation}
    where $\alpha\!=\!\frac{1}{L_{\psi}\eta-1}$ can be regarded as a hyper-parameter to balance the model performance over new and past contexts.
\end{lemma}

\section{Evaluation}
\label{sec: Evaluation}

\subsection{Experimental Setup}
\label{sec: experimental setup}
\textbf{Tasks, Datasets and Models.}
To demonstrate {\ttfamily Delta}'s broad applicability, we evaluate {\ttfamily Delta} on four typical mobile computing tasks with diverse data modalities, model structures and categories of user contexts (summarized in Table \ref{tab: summary of tasks contexts and datasets}).

\noindent $\bullet$ \textit{Image Classification (IC).} 
The Cifar10-C dataset~\cite{hendrycks2019robustness} contains around $750,000$ images of 10 objects across four context categories: weather, noise, blur and digital corruptions. For each context category, the dataset is processed into 5 subsets with 2 new objects and 1 new context per subset. ResNet-18~\cite{he2016deep} is trained for this 10-class image classification task.

\noindent $\bullet$ \textit{Human Activity Recognition (HAR).} 
HHAR~\cite{stisen2015smart}, UCI~\cite{reyes2016transition}, MotionSense~\cite{malekzadeh2019mobile} and Shoaib~\cite{shoaib2014fusion} are four public datasets collected from 73 users performing 6 basic activities (still, walking, upstairs, downstairs, jogging, bike) with 5 device placements (pocket, belt, arm, wrist, waist). For each context category, the dataset is processed into 6 subsets with 1 new activity in a new context. A lightweight CNN-based model DCNN~\cite{yang2015deep} is trained for this 6-class classification task.

\noindent $\bullet$ \textit{Audio Recognition (AR).} 
Google Speech command~\cite{warden2018speech} comprises 100,000 sound files of 20 commands from over 2,000 users with varied tones and environmental conditions. The dataset is processed into 5 subsets for each context category, each containing 4 new commands in 1 new context.
A deep neural network VGG-11~\cite{simonyan2014very} is deployed for this task.

\noindent $\bullet$ \textit{Text Classification (TC).} 
The NC corpus in XGLUE benchmark~\cite{Liang2020XGLUEAN} is a cross-lingual understanding dataset consisting of $50, 000$ articles on $10$ topics and in $5$ languages (German, English, Spanish, French, Russian). 
For each context category, the dataset is processed into 5 subsets with 2 new topics and 1 new context.
A transformer-based model BERT~\cite{devlin2018bert} is fine-tuned for this 10-class classification task.

\noindent Note that we standardize the total number of on-device contexts to approximately $5$ to ensure a consistent evaluation of {\ttfamily Delta} across various tasks, models and modalities, and thus the class number per context may vary for different datasets.

\textbf{Configurations.}
For each task, we collect data from
50$\%$ users (or randomly select 50$\%$ samples for IC and TC tasks) to form the cloud-side public dataset, with the remaining data used to simulate the on-device empirical data across different contexts. 
For cloud server, data samples from different users and contexts are mixed to reflect the typical scenario where the specific context of each raw data sample is unknown.
For mobile device, we use $5$ samples per class in each context as empirical data for model fine-tuning, consistent with the statistics that an average European citizen takes over around $4.9$ photos daily~\cite{image_scarcity} and uses Siri several times a day~\cite{siri_scarcity}. The remaining data samples are used as testing data for each context.
For {\ttfamily Delta}, the temperature $\tau$ for device-side soft matching is set to $0.1$ and the number of cloud-side data clusters is $20\!\times\!num\_class$ (i.e. $200/120/400/200$ for IC/HAR/AR/TC). The hyperparameter $\alpha$ is set to 1.0 to balance the effects of cloud-side data sampling on new and past contexts. The default communication budget is set to $25$ samples/class for each new context, and an in-depth analysis of the impacts of such budget and on-device data amount is presented in \S\ref{sec: component-wise analysis}.
\begin{table*}[]
    \centering
    \begin{small}
    \begin{tabular}{|p{0.6cm}<{\centering}|p{1.15cm}<{\centering}|p{8.45cm}<{\centering}|p{3.55cm}<{\centering}|p{2.2cm}<{\centering}|}
        \hline
        \textbf{Task} & \textbf{Modality} & \textbf{Context Category} & \textbf{Dataset} & \textbf{Model(\#params)}\\
        \hline
        IC & Image & Object (O), Weather (W), Noise (N), Blur (B), Digital Corruption (D) & Cifar10-C & ResNet18(11.2M)\\
        \hline
        HAR & IMU & Activity (A), Physical Condition (P), Device Placement (D) & HHAR, UCI, Motion, Shoaib & DCNN(17.3K)\\
        \hline
        AR & Audio & User Command (C), Tone (T), Environmental Noise (N) & Google Speech & VGG11(9.75M)\\
        \hline
        TC & Text & Article Topic (T), Language (L) & XGLUE & BERT(0.178B)\\
        \hline
    \end{tabular}
    \end{small}
    \vspace{-0.1cm}
    \caption{Summary of tasks, modalities, contexts, datasets and models.}
    \vspace{-0.4cm}
    \label{tab: summary of tasks contexts and datasets}
\end{table*}

\textbf{Baselines.}
To our best knowledge, {\ttfamily Delta} is the first data enrichment framework for on-device CL, and we compare it against the model- and algorithm-based baselines (few-shot CL and federated CL) and a random data enrichment baseline.
1) \textit{Few-shot CL} pre-trains model on cloud-side data in advance to capture the general knowledge, which is transferred to device-side new contexts through knowledge distillation~\cite{zhao2023few} ({\ttfamily FS-KD}), robust optimization~\cite{shi2021overcoming} ({\ttfamily FS-RO}) and parameter freezing~\cite{mazumder2021few} ({\ttfamily FS-PR}). 
2) \textit{Federated CL} leverages the cloud server to periodically aggregate the models trained on multiple devices per $10$ local model updates. In our experiments, the default device number is 50, except for 35 for HAR task. We use {\ttfamily Fed-$p$} to denote different settings of device participation rate $p$. For IC and TC tasks, the data samples from each user are from the same context category to simulate the common data heterogeneity across users (e.g.  W/N/B/D for IC and L for TC). 
The CL performance is evaluated on an independent test dataset, constructed according to the user contexts specified by the experimental setting.
3) \textit{Random} method selects a random cloud-side data-subset to enrich device-side empirical data.

\textbf{Metrics.}
We assess the on-device CL performance using four metrics.
\textit{Overall performance} measures the inference accuracy of the final model across all the encountered contexts. 
\textit{Learning plasticity} is the average of each new context's highest accuracy during its learning process.
\textit{Memory stability} is the average ratio between each context's final accuracy to its maximal accuracy.
\textit{System overheads} include the computation latency, communication costs, memory footprint and energy consumption for both the device side and cloud side.

\textbf{Deployments.}
We use a cloud server with one NVIDIA 3090Ti GPU and one mobile platform NVIDIA Jetson Nano~\cite{nano}.  

\subsection{End-to-End Performance}
We begin by comparing the end-to-end performance of {\ttfamily Delta} against the baselines across all four tasks. 
\begin{table*}[]
    \centering
    \begin{footnotesize}
    \begin{tabular}{|c|c|c|ccc|ccc|cc|c|p{1cm}<{\centering}|}
        \hline
        \multirow{2}{*}{\textbf{Tasks}} & \textbf{Context} & \textbf{Vanilla} & \multicolumn{3}{c|}{\textbf{Few-Shot CL}} & \multicolumn{3}{c|}{\textbf{Federated CL}} & \multicolumn{2}{c|}{\textbf{Data Enrichment}} & \multirow{2}{*}{\textbf{$\Delta$Acc.}} & \multirow{2}{*}{\textbf{$\Delta$Comm.}}\\
        \cline{4-11}
        & \textbf{Category} & \textbf{CL} & FS-KD & FS-RO & FS-PF & Fed-0.1 & Fed-0.2 & Fed-0.4 & Random & Delta & & \\
        \hline
        \multirow{5}{*}{IC} & O+W & 32.7{\tiny $\pm1.49$} & 41.7{\tiny $\pm1.78$} & 39.2{\tiny $\pm2.13$} & 36.9{\tiny $\pm2.87$} & 31.8{\tiny $\pm0.24$} & 46.4{\tiny $\pm1.65$} & 55.1{\tiny $\pm0.42$} & 42.5{\tiny $\pm2.42$} & \cellcolor{gray!25}57.7{\tiny $\pm0.54$} & $16.0\%\uparrow$ & $93.7\%\downarrow$\\
        & O+N & 31.3{\tiny $\pm1.74$} & 36.2{\tiny $\pm2.34$} & 35.5{\tiny $\pm1.65$} & 32.3{\tiny $\pm1.25$} & 31.1{\tiny $\pm0.04$} & 40.4{\tiny $\pm0.51$} & 45.0{\tiny $\pm0.12$} & 35.8{\tiny $\pm1.00$} & \cellcolor{gray!25}50.9{\tiny $\pm1.66$} & $14.8\%\uparrow$ & $93.5\%\downarrow$\\
        & O+B & 35.6{\tiny $\pm0.94$} & 43.7{\tiny $\pm1.12$} & 40.6{\tiny $\pm0.24$} & 39.2{\tiny $\pm0.06$} & 32.6{\tiny $\pm0.16$} & 39.6{\tiny $\pm0.24$} & 50.1{\tiny $\pm0.31$} & 39.9{\tiny $\pm1.69$} & \cellcolor{gray!25}57.7{\tiny $\pm0.98$} & $14.0\%\uparrow$ & $91.1\%\downarrow$\\
        & O+D & 45.0{\tiny $\pm2.57$} & 55.1{\tiny $\pm1.17$} & 51.5{\tiny $\pm2.66$} & 52.2{\tiny $\pm3.10$} & 36.9{\tiny $\pm0.04$} & 49.0{\tiny $\pm0.51$} & 61.7{\tiny $\pm0.34$} & 53.7{\tiny $\pm2.24$} & \cellcolor{gray!25}72.3{\tiny $\pm2.27$} & $17.1\%\uparrow$ & $92.2\%\downarrow$\\
        & O+W+N+B+D & 77.3{\tiny $\pm0.49$} & 81.2{\tiny $\pm1.53$} & 80.4{\tiny $\pm0.81$} & 75.3{\tiny $\pm0.41$} & 30.0{\tiny $\pm0.05$} & 39.8{\tiny $\pm0.71$} & 50.8{\tiny $\pm0.41$} & 47.8{\tiny $\pm6.64$} & \cellcolor{gray!25}94.8{\tiny $\pm2.74$} & $13.6\%\uparrow$  & $95.3\%\downarrow$\\
        \hline
        \multirow{3}{*}{HAR} & A & 52.4{\tiny $\pm3.67$} & 55.0{\tiny $\pm3.93$} & 52.9{\tiny $\pm2.55$} & 48.3{\tiny $\pm2.69$} & 54.0{\tiny $\pm0.64$}& 60.0{\tiny $\pm0.21$} & 61.3{\tiny $\pm0.55$} & 58.4{\tiny $\pm0.35$} & \cellcolor{gray!25}69.3{\tiny $\pm1.96$} & $14.3\%\uparrow$  & $99.6\%\downarrow$\\
        & A+P & 51.2{\tiny $\pm4.53$} & 53.3{\tiny $\pm3.20$} & 50.1{\tiny $\pm3.52$} & 49.4{\tiny $\pm2.95$} & 60.5{\tiny $\pm1.28$} & 61.1{\tiny $\pm1.89$} & 63.1{\tiny $\pm0.85$} & 58.5{\tiny $\pm0.75$} & \cellcolor{gray!25}66.6{\tiny $\pm1.78$} & $13.3\%\uparrow$ & $99.8\%\downarrow$\\
        & A+P+D & 81.0{\tiny $\pm4.75$} & 80.3{\tiny $\pm2.35$} & 78.7{\tiny $\pm4.37$} & 71.0{\tiny $\pm4.27$} & 62.2{\tiny $\pm3.58$} & 66.8{\tiny $\pm3.97$} & 70.1{\tiny $\pm4.28$} & 61.1{\tiny $\pm3.25$} & \cellcolor{gray!25}90.3{\tiny $\pm5.09$} & $10.0\%\uparrow$ &  $99.7\%\downarrow$\\
        \hline
        \multirow{3}{*}{AR} & C & 93.6{\tiny $\pm0.16$} & 93.5{\tiny $\pm0.07$} & 92.9{\tiny $\pm0.65$} & 94.2{\tiny $\pm0.28$} & 88.1{\tiny $\pm1.65$} & 88.3{\tiny $\pm0.83$} & 88.5{\tiny $\pm1.78$} & 90.4{\tiny $\pm0.19$} & \cellcolor{gray!25}94.3{\tiny $\pm0.17$} & $0.2\%\uparrow$  & $99.9\%\downarrow$\\
        & C+T & 89.0{\tiny $\pm0.41$} & 89.4{\tiny $\pm0.57$} & 89.4{\tiny $\pm0.38$} & 90.3{\tiny $\pm0.79$} & 86.5{\tiny $\pm0.24$} & 88.5{\tiny $\pm0.62$} & 88.7{\tiny $\pm0.25$} & 90.3{\tiny $\pm0.26$} & \cellcolor{gray!25}91.1{\tiny $\pm1.17$} & $0.8\%\uparrow$ & $99.9\%\downarrow$\\
        & C+T+N & 84.7{\tiny $\pm0.64$} & 84.8{\tiny $\pm1.52$} & 86.2{\tiny $\pm0.79$} & 86.9{\tiny $\pm0.40$} & 87.5{\tiny $\pm0.54$} & 87.7{\tiny $\pm0.31$} & 88.0{\tiny $\pm0.61$} & 88.5{\tiny $\pm1.45$} & \cellcolor{gray!25}89.2{\tiny $\pm1.60$} & $2.3\%\uparrow$ & $99.9\%\downarrow$\\
        \hline
        \multirow{2}{*}{TC} & T & 73.2\tiny{$\pm2.15$} & 73.5\tiny{$\pm1.35$} & 75.7\tiny{$\pm4.07$} & 73.3\tiny{$\pm2.56$} & 79.6\tiny{$\pm0.37$} & 79.6\tiny{$\pm0.19$} & 79.8\tiny{$\pm0.14$} & 73.9\tiny{$\pm2.69$} & \cellcolor{gray!25}83.1\tiny{$\pm2.26$} & $7.3\%\uparrow$ & $99.8\%\downarrow$\\
        & T+L & 77.7\tiny{$\pm3.19$} & 82.2\tiny{$\pm0.29$} & 80.1\tiny{$\pm3.02$} & 80.0\tiny{$\pm1.89$} & 84.3\tiny{$\pm0.14$} & 84.4\tiny{$\pm0.18$} & 84.7\tiny{$\pm0.09$} & 79.7\tiny{$\pm2.21$} & \cellcolor{gray!25}86.2\tiny{$\pm$2.16} & $4.0\%\uparrow$ & $99.4\%\downarrow$\\
        \hline
    \end{tabular}
    \end{footnotesize}
    \caption{Summary of overall CL performance (average accuracy of final model on all contexts). We also mark {\ttfamily Delta}'s improvement on accuracy (over few-shot CL) and reduction in communication costs (over federated CL).}
    \vspace{-0.3cm}
    \label{tab: overall performance}
\end{table*}

\textbf{{\ttfamily Delta} significantly improves the overall performance of on-device CL.} 
Table \ref{tab: overall performance} summarizes the average accuracy of the final model across all contexts. 
Compared with the best-performing few-shot CL method, {\ttfamily Delta} achieves a notable improvement, with accuracy increases of $13\!-\!16\%$ higher accuracy on IC, $10\!-\!14\%$ on HAR, $0.2\!-\!2.5\%$ on AR, and $4\!-\!7.3\%$ on TC. 
Note that {\ttfamily Delta}'s improvement on AR task is minimal because its data heterogeneity across contexts is relatively low (i.e. different tones and background noises) and vanilla CL could perform well.
When compared to federated CL, {\ttfamily Delta} consistently achieves the highest overall performance across all settings, and reduces total communication costs by $91\!-\!99\%$, demonstrating its superior effectiveness and efficiency in enhancing CL performance.
Furthermore, we observe that for most tasks (IC, HAR and TC), all methods tend to perform better on contexts with mixed categories (last line of each task in Table \ref{tab: overall performance}). The potential reason is that data samples with different context categories exhibit a greater distribution divergence, making it easier for the on-device model to learn the decision boundary. 

\begin{figure*}
    \centering
    \vspace{-0.3cm}
    \includegraphics[width=0.67\textwidth]{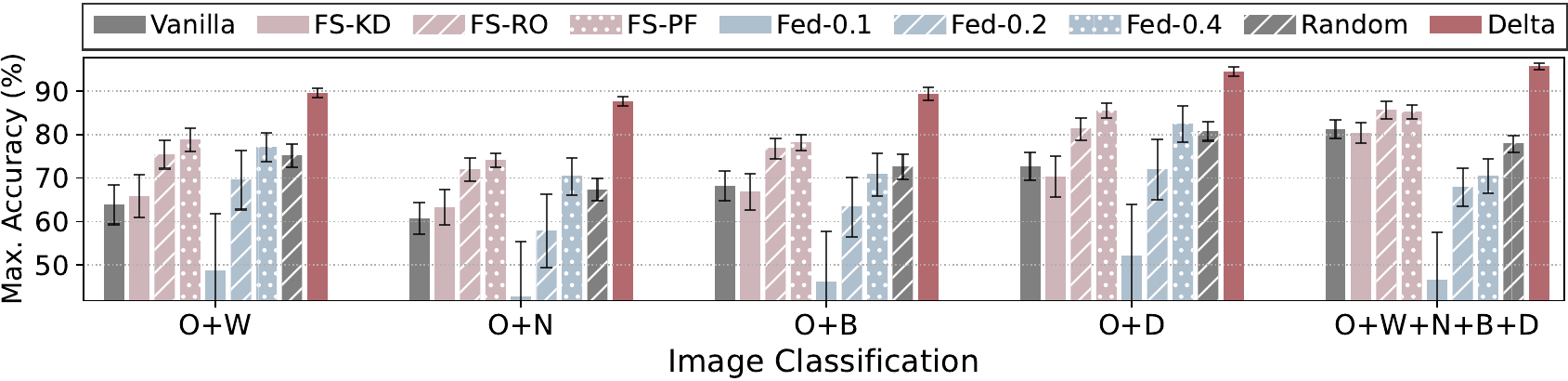}
    \includegraphics[width=0.29\textwidth]{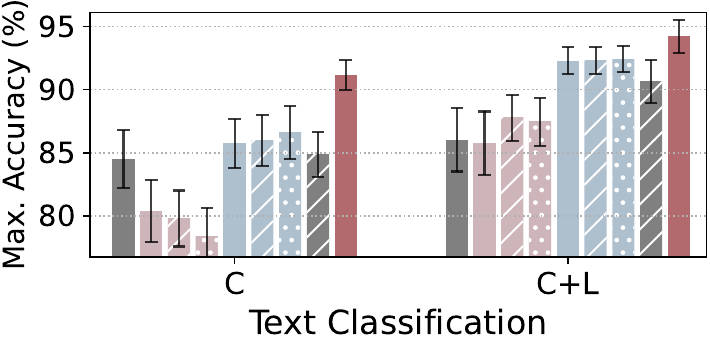}
    \includegraphics[width=0.43\textwidth]{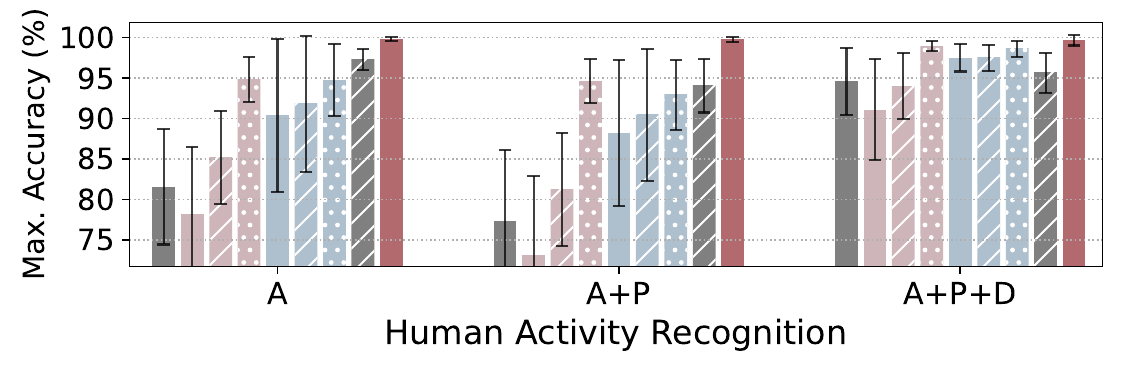}
    \quad\quad\quad\quad\quad\ 
    \includegraphics[width=0.43\textwidth]{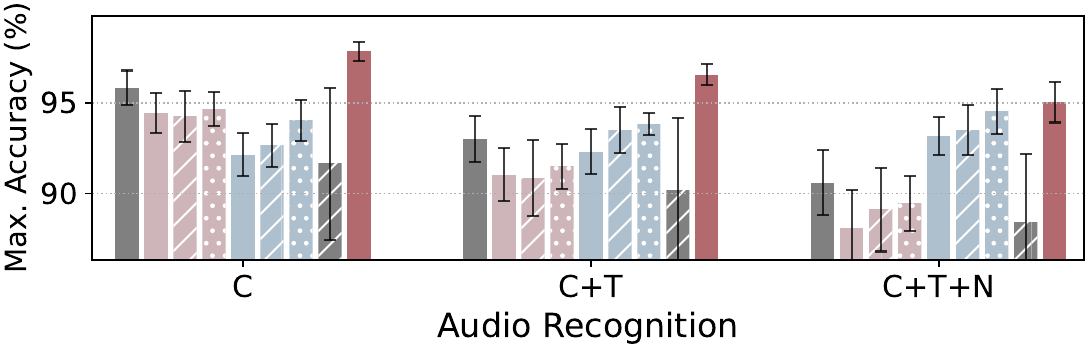}
    \vspace{-0.35cm}
    \caption{Comparison of learning plasticity (maximum model accuracy for each new context during CL).}
    \Description{Comparison of learning plasticity (maximum model accuracy for each new context during CL).}
    \vspace{-0.35cm}
    \label{fig: learning plasticity}
\end{figure*}

\textbf{{\ttfamily Delta} enhances the learning plasticity of on-device CL with various new contexts.} Figure \ref{fig: learning plasticity} reports the average value of each new context's peak accuracy during the learning process, a metric widely adopted to assess the learning plasticity.
A key observation is that {\ttfamily Delta} consistently outperforms the baselines across various tasks, data modalities and context categories, demonstrating high robustness and applicability to diverse new contexts.
For example, {\ttfamily Delta} achieves around $90\%$ and $100\%$ accuracy for new context in IC and HAR tasks regardless of context categories and fluctuates less than $3\%$ accuracy on the other two tasks. The high accuracy for new contexts can be attributed to the limited classes within each new context and the enriched data from cloud side.
In contrast, the performance of baselines on new contexts is sensitive to the diversity of context categories, such as few-shot CL dropping from $95\%$ to $90\%$ on AR task and federated CL reducing from $93\%$ to $87\%$ on TC task. 
The rationale behind these is that few-shot CL depends on the high relevance between the on-device context and the base contexts during pre-training to facilitate effective knowledge transfer. Similarly, the performance of federated CL is largely influenced by the data heterogeneity across different users' ongoing contexts.
Conversely, {\ttfamily Delta} can consistently identify an appropriate cloud-side data-subset that contributes to the device-side CL process, making it relatively robust.

\begin{figure}
    \centering
    \vspace{-0.1cm}
    \includegraphics[width=0.32\textwidth]{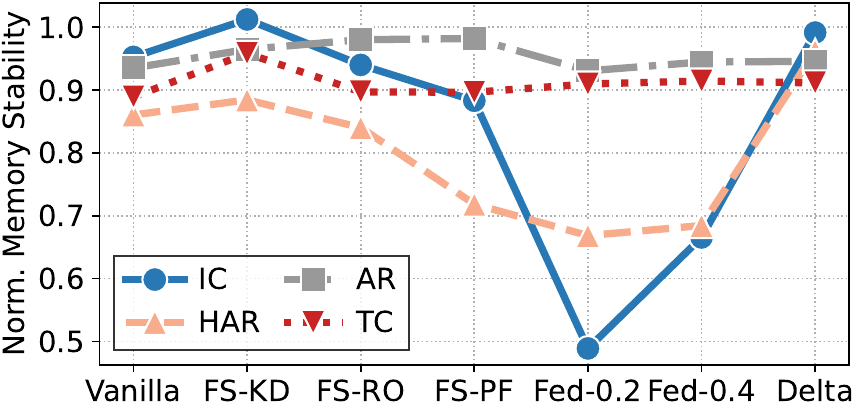}
    \vspace{-0.35cm}
    \caption{Comparison of memory stability in the settings of mixed context categories for each task.}
    \Description{Comparison of memory stability in the settings of mixed context categories for each task.}
    \vspace{-0.35cm}
    \label{fig: memory stability}
\end{figure}

\textbf{{\ttfamily Delta} consistently achieves a low accuracy drop on past contexts and exhibits high memory stability.} Figure \ref{fig: memory stability} plots the average ratio between each context's final accuracy and its peak accuracy, which indicates that {\ttfamily Delta} can maintain over $90\%$ relative performance for past contexts. The superior memory stability is due to the consideration of the impact of new context's enriched on all contexts' overall performance during the cloud-side data sampling process. 
We also observe that few-shot CL methods can slightly outperform {\ttfamily Delta} in some cases. This is because they achieve significantly lower peak accuracy for new contexts compared to {\ttfamily Delta} (e.g. a $10\%$ accuracy gap in IC task shown in Figure \ref{fig: learning plasticity}), and thus the accuracy drop might be less pronounced.

\begin{figure*}
    \vspace{-0.25cm}
    \captionsetup[figure]{skip=-1pt} 
    \centering
    \subfigure[Latency]{
        \includegraphics[height=2.65cm]{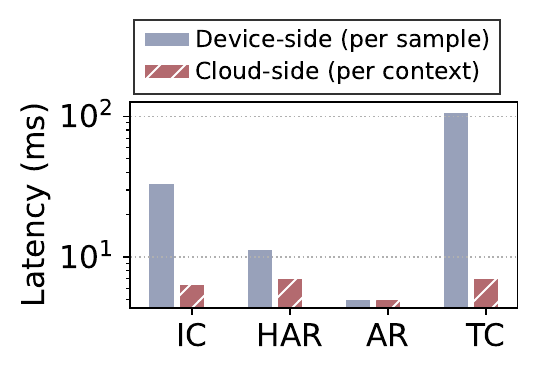}
        \label{fig: latency}
    }
    \subfigure[Energy Consumption]{
        \includegraphics[height=2.65cm]{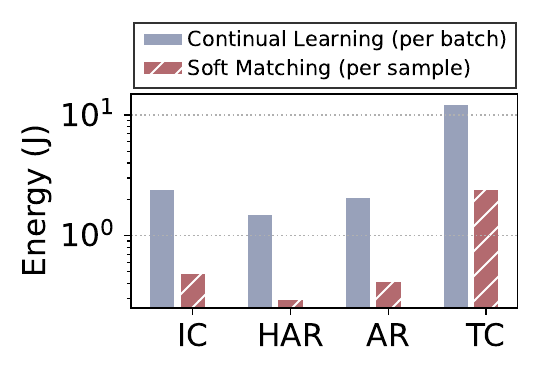}
    }
    \subfigure[Memory Footprint]{
        \includegraphics[height=2.65cm]{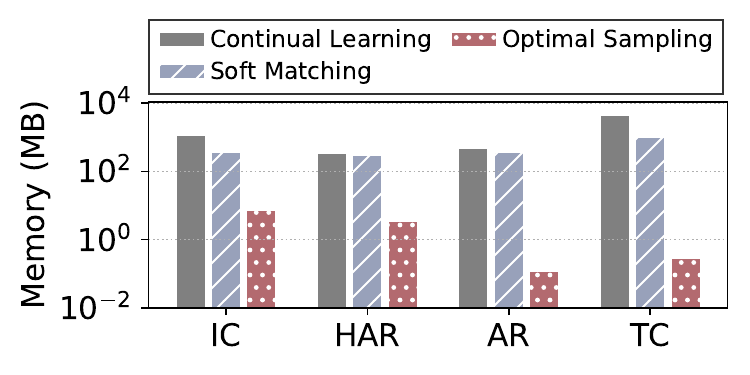}
        \label{fig: memory footprint}
    }
    \subfigure[Communication Cost]{
        \includegraphics[height=2.65cm]{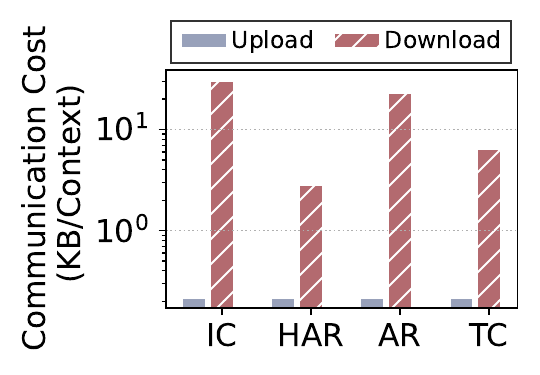}
    }
    \vspace{-0.4cm}
    \caption{System overheads of {\ttfamily Delta}.}
    \Description{System overheads of {\ttfamily Delta}.}
    \vspace{-0.15cm}
    \label{fig: system overheads}
\end{figure*}

\textbf{{\ttfamily Delta} incurs marginal system overheads for both mobile device and cloud server}, as depicted in Figure \ref{fig: system overheads}.

\noindent $\bullet$ \textit{Device-Side.} The soft matching solution for sub-problem (2a) requires computing the feature of each local data sample and its distance to each element of directory dataset. This results in additional latency of 23.8/1.05/ 4.25/109ms and energy consumption of $0.49/0.30/0.42/2.47$J per sample for IC/HAR/AR/TC tasks, respectively.
Moreover, Figure (\ref{fig: memory footprint}) shows that soft matching process has a lower memory footprint than CL process for avoiding model backpropagation, indicating that {\ttfamily Delta} does not increase peak memory usage due to the sequential execution of {\ttfamily Delta} and on-device CL.

\noindent $\bullet$ \textit{Cloud-Side.}
The analytical solution for optimal cloud-side data sampling can be computed within $2.56\!-\!7.15$ ms using a single 10-core Intel CPU with a memory footprint of $0.12\!-\!7.8$ MB. This high computational efficiency allows for parallel cloud-side operations for numerous devices simultaneously.

\noindent $\bullet$ \textit{Device-cloud Communication.}
For each context, the communication overhead includes the uploading of device-side directory weight, which consists of only several vectors ($\le1$KB), and the downloading of cloud-side enriched data, which requires a total of 30.4/2.89/23.5/6.43 KB for IC/HAR/ AR/TC tasks under default settings.

\subsection{Component-Wise Analysis}
\label{sec: component-wise analysis}

\begin{figure*}
    \vspace{-0.1cm}
    \captionsetup[subfigure]{skip=-0.5cm} 
    \begin{minipage}[b]{0.38\linewidth}
        \centering
        \subfigure[Soft Matching Temperature]{
            \includegraphics[width=0.85\textwidth]{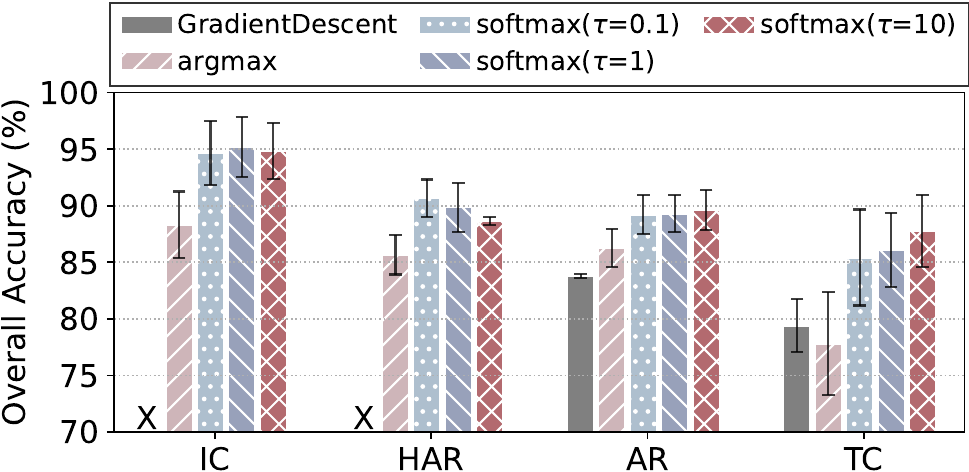}
            \label{fig: impact of soft matching}
        }
        \vspace{-0.2cm}
        \subfigure[On-Device Data Size]{
            \includegraphics[width=\textwidth]{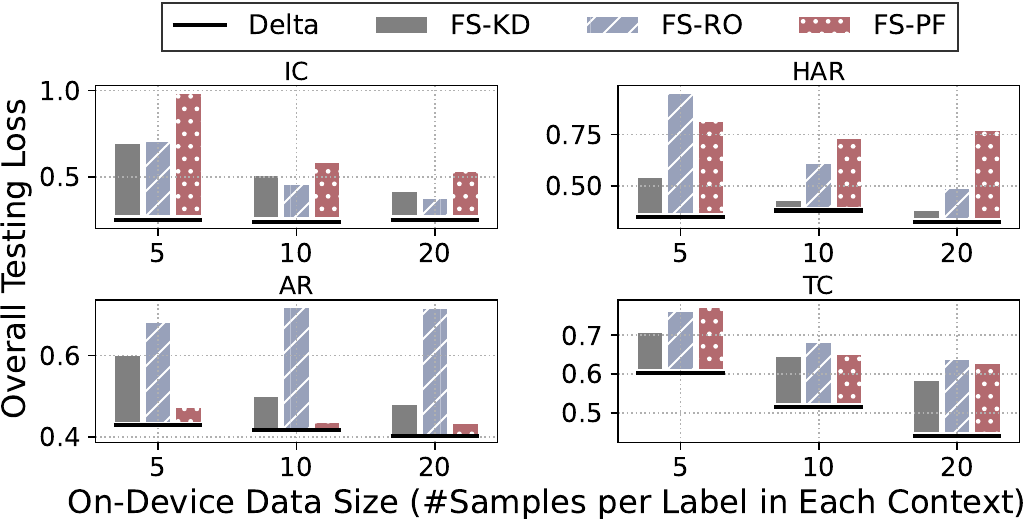}
            \label{fig: impact of data size}
        }
        \vspace{-0.3cm}
        \caption{Device-side Sensitivity Analysis.}
        \Description{Device-side Sensitivity Analysis.}
        \vspace{-0.35cm}
    \end{minipage}
    \ \ \ 
    \begin{minipage}[b]{0.38\linewidth}
        \centering
        \subfigure[Optimal Data Sampling]{
            \includegraphics[width=0.85\textwidth]{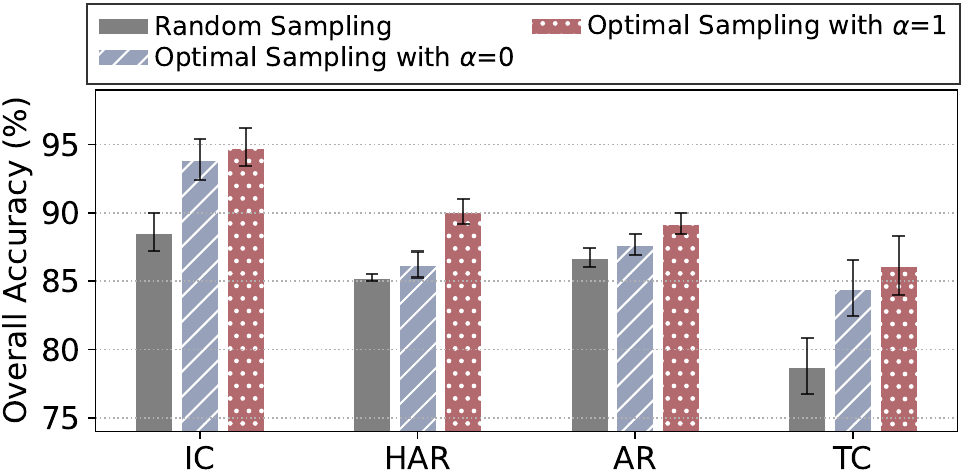}
            \label{fig: impact of optimal sampling}
        }
        \vspace{-0.2cm}
        \subfigure[Cluster Number of Directory Dataset]{
            \includegraphics[width=\textwidth]{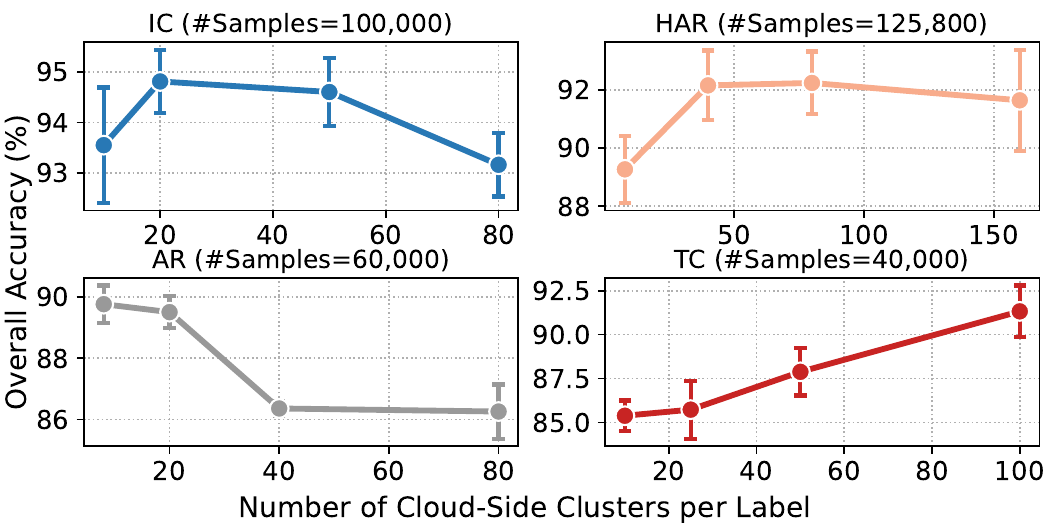}
            \label{fig: impact of cluster num}
        }
        \vspace{-0.3cm}
        \caption{Cloud-side Sensitivity Analysis.}
        \Description{Cloud-side Sensitivity Analysis.}
        \vspace{-0.35cm}
    \end{minipage}
    \ \ \ 
    \begin{minipage}[b]{0.2\linewidth}
        \centering
        \includegraphics[width=0.85\textwidth]{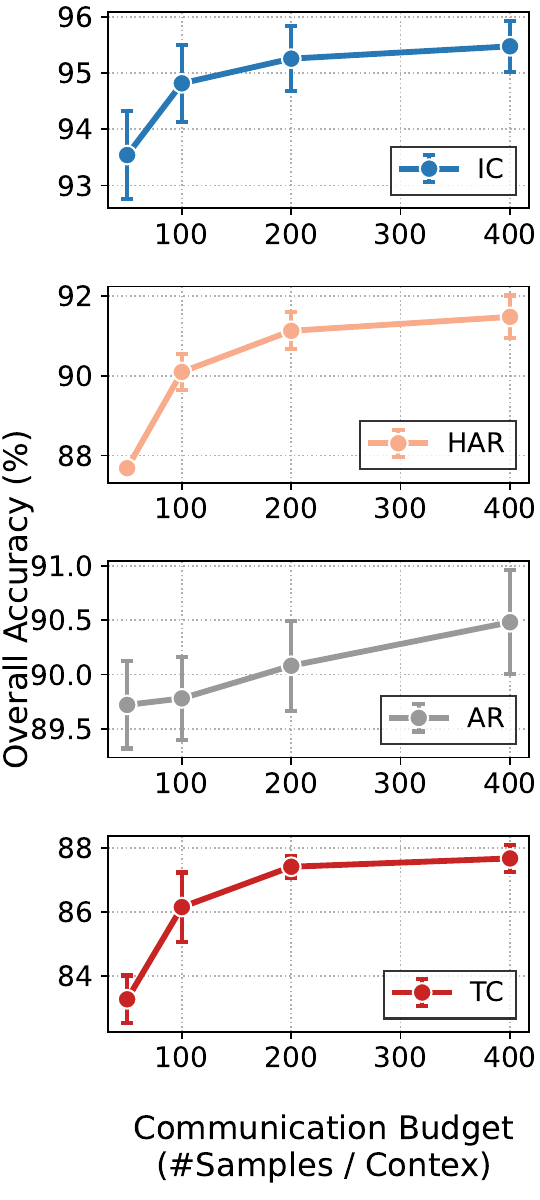}
        \vspace{-0.4cm}
        \caption{Impact of Cloud-Device Communication Budget.}
        \Description{Impact of Cloud-Device Communication Budget.}
        \vspace{-0.35cm}
        \label{fig: impact of acqu num}
    \end{minipage}
\end{figure*}

We further delve into the functionality and sensitivity of each key component within {\ttfamily Delta} framework.

\textbf{Device-Side Data Soft Matching.}
To illustrate the importance of soft matching strategy, we assess the performance of {\ttfamily Delta} using various strategies to address sub-objective (2a), including gradient descent ($GD$), hard matching ($argmax$) and soft matching ($softmax$) with varying temperatures $\tau$. Figure \ref{fig: impact of soft matching} indicates that $GD$ underperforms across most tasks, while $softmax$ consistently outperforms $argmax$. 
The reasons are twofold: 
1) $GD$ is susceptible to getting trapped in local optima and leads to the overfitted directory weight; 
2) $argmax$ fails to exploit the similarities between one device-side sample and multiple cloud-side clusters, which is essential when the cloud-side data is finely clustered.
We also note that the optimal $\tau$ differs by task due to varying feature distributions, and we set $\tau\!=\!1.0$ for stable performance.

\textbf{Device-Side Data Size.}
Figure \ref{fig: impact of data size} shows the impact of user data amount on {\ttfamily Delta} and baseline performances, where we present testing loss instead of accuracy for clearer comparison.
{\ttfamily Delta} demonstrates relatively high robustness, which is attributed to
1) the effective solution of the device-side sub-problem with scarce on-device data through our soft matching strategy, and
2) the substantial performance improvement brought by the abundant cloud-side enriched data compared to additional device-side user data.
Also, we observe that baselines show greater sensitivity to on-device data quantity, highlighting the critical role of on-device data enrichment and further motivates our work.

\textbf{Cloud-Side Directory Dataset.}
Figure \ref{fig: impact of cluster num} plots the performance of {\ttfamily Delta} with varying numbers of cloud-side data clusters for directory dataset construction, where we replace the cloud-side data sampling scheme with random sampling to isolate the effects of directory dataset. We observe that a slight increase in cluster number can improve {\ttfamily Delta}'s performance by making directory dataset more representative and aligning the cloud-side sub-objective (2b) more closely with the overall objective (\ref{eq: data enrichment problem}). However, an excessively large cluster number can result in numerous similar clusters, leading to the selection of redundant data for enrichment. For stable performance, we set cluster number per label to 20.

\textbf{Cloud-Side Optimal Data Sampling.}
To evaluate the importance of cloud-side optimal data sampling, we assess {\ttfamily Delta}'s performance with different sampling schemes, including random sampling, optimal sampling for solely new context ($\alpha\!=\!0$) and optimal sampling considering all contexts ($\alpha\!=\!1$). 
Figure \ref{fig: impact of optimal sampling} indicates that optimal data sampling for only new context improves overall model accuracy by $5.3/0.9/1.0/5.7\%$ for IC/HAR/AR/TC tasks. Considering past contexts further enhances accuracy by $0.9/3.9/1.5/1.7\%$. 
Notably, he most significant improvements are observed in IC and TC tasks, as the visual and textual data we used are more diverse, making random sampling less stable and effective.

\textbf{Device-Cloud Communication Budget.}
We further evaluate {\ttfamily Delta}'s performance with varying sizes of cloud-side enriched data to simulate different communication budgets. Figure \ref{fig: impact of acqu num} shows that {\ttfamily Delta}'s performance improves significantly as the enriched data size per context increases from $50$ to $100$, and then stabilizes with larger data sizes. This robustness highlights {\ttfamily Delta}'s  applicability for real-world devices with diverse network conditions.

\section{Related Work}
\label{sec: Related Work}

\textbf{Cloud-Side Continual Learning}
aims to train ML models over non-stationary data streams to acquire new contextual knowledge without forgetting past contexts. 
This approach is inspired by the capability of biological neural networks to modulate synaptic memory and plasticity in response to dynamic inputs~\cite{zenke2017continual, wang2024comprehensive}.
Existing solutions include: 
1) stabilizing previously-learned synaptic changes by penalizing parameter changes of the past optimal model~\cite{kirkpatrick2017overcoming, zenke2017continual};
2) expanding and pruning synaptic connections to form new synaptic memory via creating additional parameter space for new contexts and re-normalizing them with past contexts~\cite{mallya2018piggyback, serra2018overcoming};
3) consolidating synaptic memory by storing the important data of past contexts and replaying them during learning new contexts~\cite{li2017learning, lopez2017gradient}, where the data importance can be measured by representativeness~\cite{rebuffi2017icarl, shim2021online}, diversity~\cite{DBLP:conf/iclr/YoonMYH22} or uncertainty~\cite{aljundi2019online}.
Previous studies~\cite{kwon2021exploring, lee2022carm, hayes2022online} have found that data replay methods provides the best trade-off between model performance and system efficiency, and thus our experiments are mainly conducted in this case. 
\textit{{\ttfamily Delta} framework serves as a plug-in component to enrich on-device data and enhance performance for all these methods.}

\textbf{Device-Side Continual Learning}
focuses on optimizing the utilization of hardware resources to implement cloud-side CL algorithms on resource-constrained devices. This includes saving storage cost through data quantization techniques~\cite{ravaglia2021tinyml, hersche2022constrained}, 
reducing memory overhead through context-aware parameter sparsity~\cite{xie2020kraken, kwon2023tinytrain}, 
accelerating data loading via hierarchical memory management~\cite{lee2022carm, ma2023cost}, 
and accelerating computation with adaptive computing resource~\cite{leite2022resource, kwon2023lifelearner, tian2018continuum, kudithipudi2023design}.
However, most of these works overlook the data bottleneck on mobile device (scarce, personal and unpredictable user data), and thus \textit{{\ttfamily Delta} is complementary to existing on-device CL works focusing on hardware bottleneck.}

\textbf{On-Device Data Augmentation}
is a powerful technique to improve model training performance by generating diverse data from existing user data, such as leveraging geometric and color space transformation and random erasing for visual images~\cite{shorten2019survey},
using techniques grounded in physical principles for IMU signal~\cite{xu2023practically}, as well as
employing language rule-based transformations and synonym replacement for textual data~\cite{bayer2022survey}.
However, a significant limitation of data augmentation is that each data modality and task necessitates specifically designed augmentation techniques  to accommodate unique data characteristics, making the data augmentation process cumbersome and inefficient. 
\textit{{\ttfamily Delta} serves as a generally solution to complement these works by directly expanding the on-device available data.}
\section{Discussion}
\label{sec: limitation and future work}

\textbf{Privacy Consideration.}
In {\ttfamily Delta} framework, the information uploaded by devices includes the directory weights, which excludes any raw user data and protects privacy like FL~\cite{mcmahan2017communication}. 
Unlike FL, where the transmitted model updates inherently encode specific features of training data, {\ttfamily Delta}'s transmitted weights only indicate the similarity between user data and directory dataset (e.g. likelihood of weather conditions rather than pixels in IC task, probability of device placement rather than specific IMU signals in HAR task), which reveals rough context information and makes the recovery or identification of raw data more challenging. 
To further enhance privacy, secure aggregation techniques like secure multi-party computation~\cite{goldreich1998secure} and homomorphic encryption~\cite{acar2018survey} can be integrated into the communication and computation processes in {\ttfamily Delta}.

\textbf{Comparison with FL.}
The intuitions behind {\ttfamily Delta} framework and FL paradigm are distinct. 
FL aims to leverage device-side data to develop a global model that can generalize well across diverse user contexts, i.e. \textit{global knowledge aggregation}.
In contrast, {\ttfamily Delta} utilizes cloud-side data to enhance the personalization of local models for individual user contexts, i.e. \textit{local knowledge augmentation}.
As a result, the applicability of FL is primarily limited by device-side constraints, including the vast number of devices, high participation rates, cross-device data heterogeneity and tolerance for communication overheads. 
{\ttfamily Delta}, on the other hand, seeks to shift the limitations to the cloud, assuming that cloud server can collect abundant public data to match different users. This aligns with the recent success of training billion-scale models over sufficiently diverse datasets for various tasks. Additionally, when confronted with extremely rare user contexts, Delta could still identify the most helpful and relevant cloud-side data-subsets to provide data foundation for existing model or algorithm-based augmentation methods.
In conclusion, FL and {\ttfamily Delta} are applicable for different scenarios and could potentially be complementary.

\section{Conclusion}
\label{section: conclusion}
In this work, we explore the potential of leveraging cloud-side abundant data resource to address the data bottleneck in on-device CL. We formalize the data enrichment problem and propose {\ttfamily Delta}, a private, efficient and effective cloud-assisted data enrichment framework for on-device CL. On extensive experiments, {\tt Delta} shows superior model performance and system efficiency across various mobile computing tasks, data modalities and model structures.

\begin{acks}
    This work was supported in part by National Key R\&D Program of China (No. 2022ZD0119100), in part by China NSF grant No. 62322206, 62132018, U2268204, 62025204, 62272307, 62372296. The opinions, findings, conclusions, and recommendations expressed in this paper are those of the authors and do not necessarily reflect the views of the funding agencies or the government.
    The authors thank the anonymous reviewers and the shepherd for their insightful feedbacks.
\end{acks}

\clearpage
\balance
\bibliographystyle{ACM-Reference-Format}
\bibliography{main}

\end{document}